\documentclass[preprint,12pt]{elsarticle}

\usepackage{amssymb}

\usepackage{amsmath}
\usepackage{mathtools}
\usepackage{amsthm}

\usepackage{microtype}
\usepackage{graphicx}
\usepackage{subfigure}
\usepackage{booktabs} 
\usepackage{multirow}
\usepackage{graphicx}
\usepackage{subfigure}
\usepackage{longtable}
\usepackage{bbm}
\usepackage{bm}

\usepackage{color}
\newcommand{\red}[1]{{#1}} 

\newtheorem{theorem}{Theorem}[section]

\newtheorem{corollary}[theorem]{Corollary}
\newtheorem{definition}[theorem]{Definition}
\newtheorem{assumption}[theorem]{Assumption}
\newtheorem{example}{Example}[section]
\usepackage{hyperref}

\journal{Neural Networks}

\begin{document}

\begin{frontmatter}



\title{Long-term Causal Effects Estimation via Latent Surrogates Representation Learning}

\author[inst1]{ Ruichu Cai \corref{cor1} \fnref{label2}}
\author[inst1]{Weilin Chen \fnref{label2}}
\author[inst1]{Zeqin Yang}
\author[inst2]{Shu Wan}
\author[inst3]{Chen Zheng}
\author[inst3]{Xiaoqing Yang}
\author[inst3]{Jiecheng Guo}

\cortext[cor1]{Coresponding author.}
\fntext[label2]{Equal Contribution.}

\affiliation[inst1]{organization={School of Computer Science, Guangdong University of Technology},
            city={Guangzhou},
            country={China}}
            
\affiliation[inst2]{organization={School of Computing and Augmented Intelligence, Arizona State University},
            city={Tempe},
            state={AZ},
            country={USA}}

\affiliation[inst3]{organization={Didi Chuxing},
            city={Beijing},
            country={China}}

\begin{abstract}
Estimating long-term causal effects based on short-term surrogates is a significant but challenging problem in many real-world applications such as marketing and medicine. Most existing methods estimate causal effects in an idealistic and simplistic manner - disregarding unobserved surrogates and treating all short-term outcomes as surrogates. However, such methods are not well-suited to real-world scenarios where the partially observed surrogates are mixed with the proxies of unobserved surrogates among short-term outcomes. To address this issue, we develop our flexible method called LASER to estimate long-term causal effects in a more realistic situation where the surrogates are either observed or have observed proxies. In LASER, we employ an identifiable variational autoencoder to learn the latent surrogate representation by using all the surrogate candidates without the need to distinguish observed surrogates or proxies of unobserved surrogates. With the learned representation, we further devise a theoretically guaranteed and unbiased estimation of long-term causal effects. Extensive experimental results on the real-world and semi-synthetic datasets demonstrate the effectiveness of our proposed method.
\end{abstract}




\end{frontmatter}


\section{Introduction}
\label{sec:intro}
\begin{figure}[!htp]
\begin{center}
    \includegraphics[width=1.\textwidth]{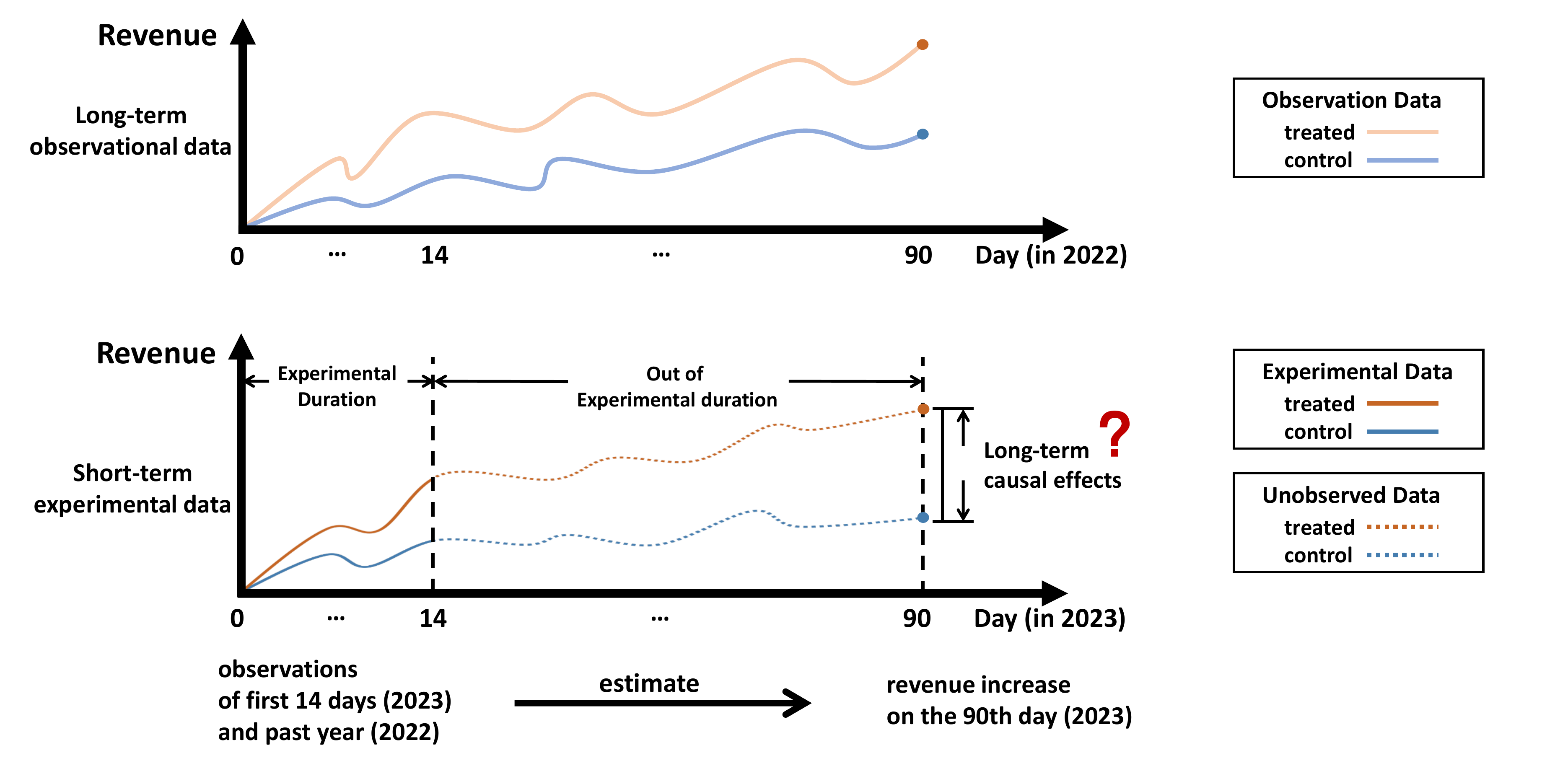}
\caption{
Example graph of long-term causal effects estimation task on a ride-hailing platform, corresponding to Example \ref{eg: intor}. \red{Our goal is to estimate long-term causal effects using short-term experimental data and long-term observational data.}
}
\label{fig:example}
\end{center}
\end{figure}

Estimating long-term causal effects is an important but challenging problem in many fields, from marketing (e.g., long-term revenue affected by high/low quality of advertising \cite{hohnhold2015focusing}), to medicine (e.g., taking years of follow-up to fully reveal mortality in \red{acquired immunodeficiency syndrome} drug clinical trials \cite{fleming1994surrogate}). A key challenge in estimating long-term causal effects is the expense of conducting a long-time experiment to obtain a long-term outcome. Existing methods \cite{prentice1989surrogate,frangakis2002principal,athey2019surrogate, cheng2021long} try to solve this problem by conducting short-time experiments and utilizing short-term outcomes as \textit{surrogates} to estimate the long-time outcomes, which is much cheaper and more feasible, as shown in Figure \ref{fig:example} and Example \ref{eg: intor}.

\begin{example} \label{eg: intor}

In the example illustrated in Figure \ref{fig:example}, to increase customer loyalty and revenue in the next quarter of 2023, a ride-hailing platform needs to evaluate long-term campaign strategies such as a 90-day taxi promotion or advertising strategy. There are two ways to estimate the long-term effects of the promotion on revenue. One way is to conduct a 90-day random control experiment, and another way is to conduct an only-14-days experiment and to collect historical 90-day data available on the platform in the previous year. Comparing the two ways, the first will result in unaffordable marketing costs, while the second is considerably more feasible and acceptable to the ride-hailing platform. 
\end{example}

Existing methods try to estimate long-term causal effects of treatment by utilizing surrogates, e.g., predicting the 90-day revenue increase of a promotion strategy by using users' first 14-day usage frequency in Example \ref{eg: intor}. 
Athey et~al. \cite{athey2019surrogate} propose Surrogate Index (SInd), a long-term outcome predictor given surrogates and covariates to estimate long-term causal effects based on \red{the} surrogacy assumption. 
Kallus \red{et~al}. \cite{kallus2020role} propose an estimator based on flexible machine learning methods to estimate nuisance parameters. 
Cheng et~al. \cite{cheng2021long} concentrate on time-dependent surrogates and employ an \red{Recurrent Neural Network (}RNN\red{)} with double heads to capture nonlinearity between surrogates and long-term outcomes. 
By utilizing the surrogates' information related to the long-term outcome in observational data, these methods propose different types of estimators for long-term causal effects.

\begin{figure}[h]
\begin{center}
\subfigure[Assumed causal graph in existing works.]{ \label{figure 1a}
    \includegraphics[width=0.28\textwidth]{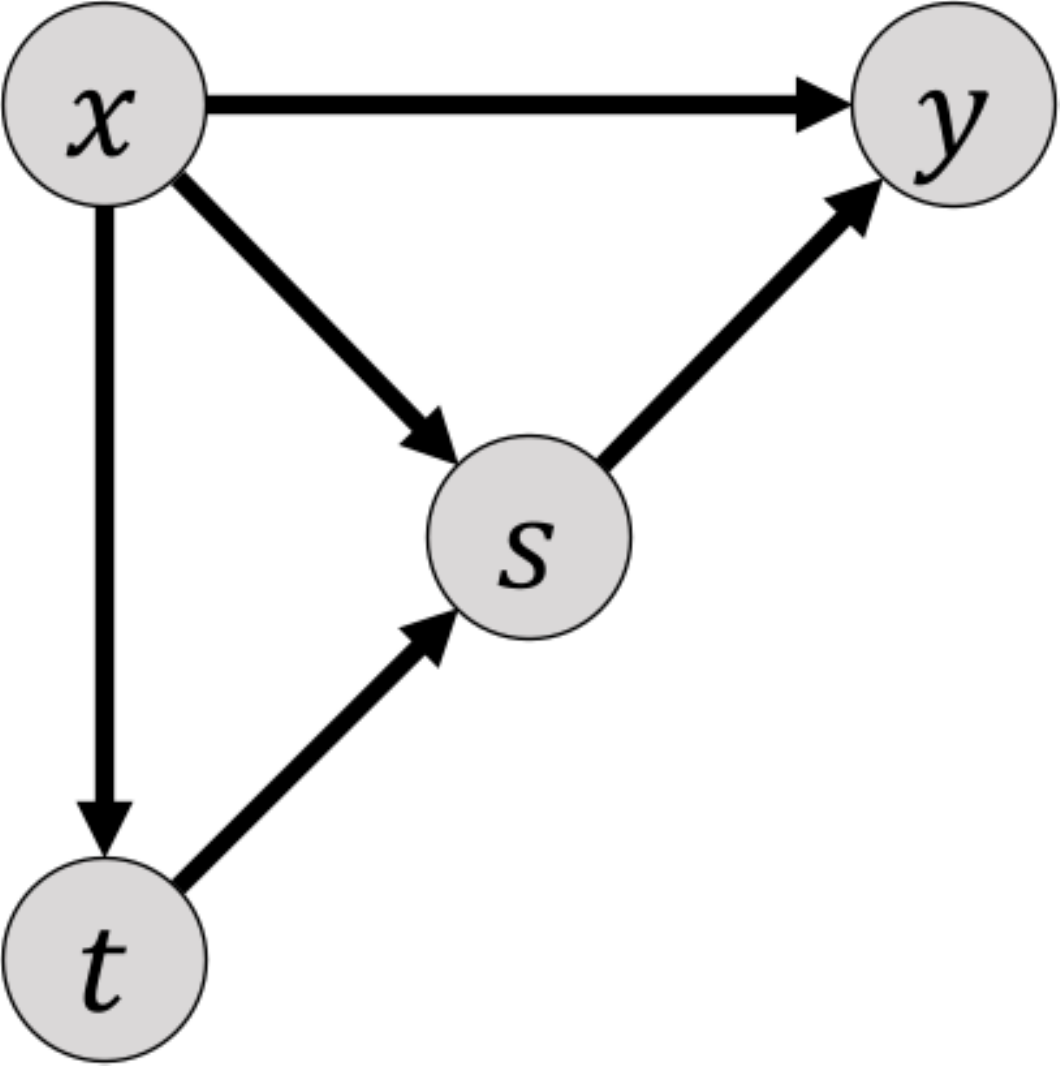}
}
\hspace{0.03\textwidth}
\subfigure[Assumed causal graph in our work.]{   \label{figure 1b}
    \includegraphics[width=0.28\textwidth]{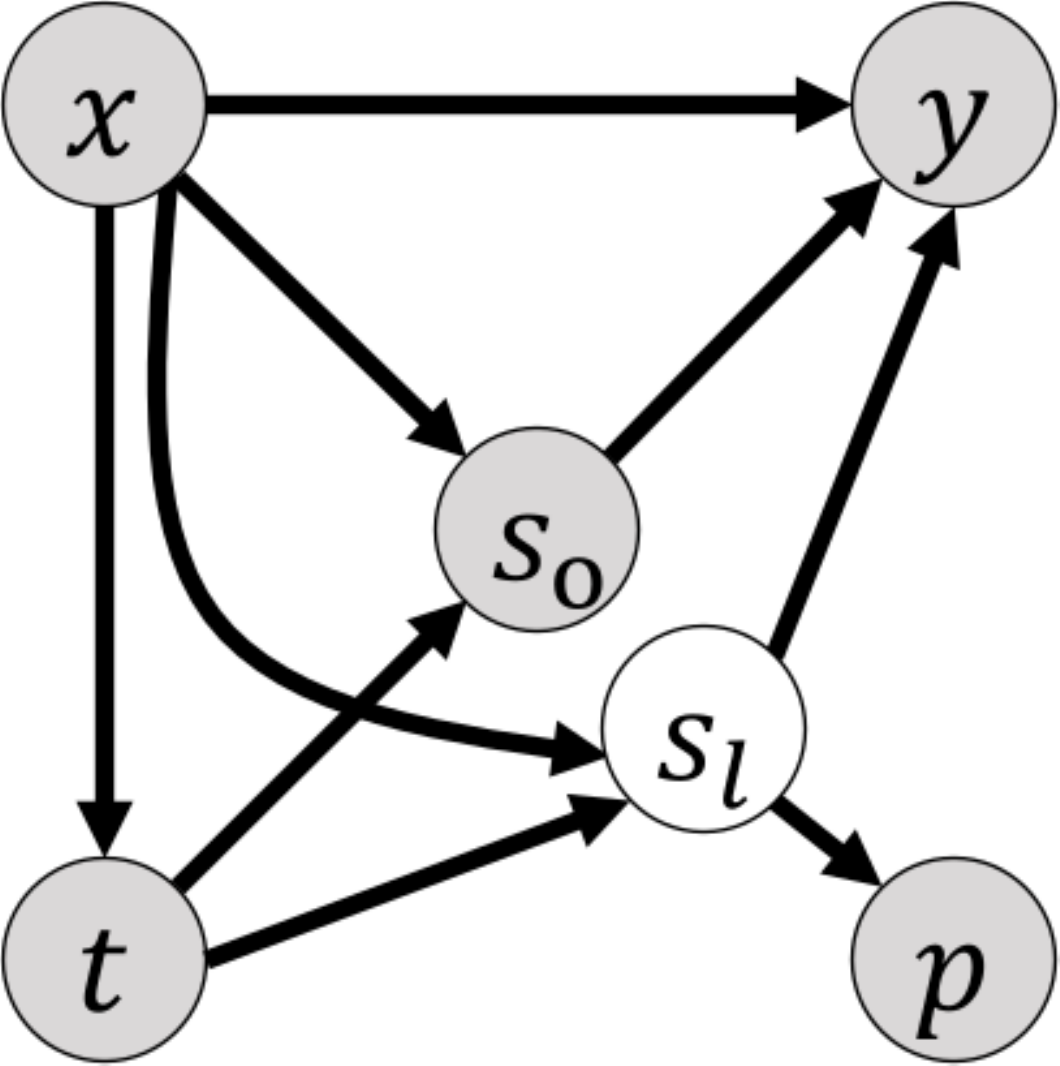}
}
\hspace{0.03\textwidth}
\subfigure[Corresponding example graph.]{   \label{figure causal graph example}
    \includegraphics[width=0.28\textwidth]{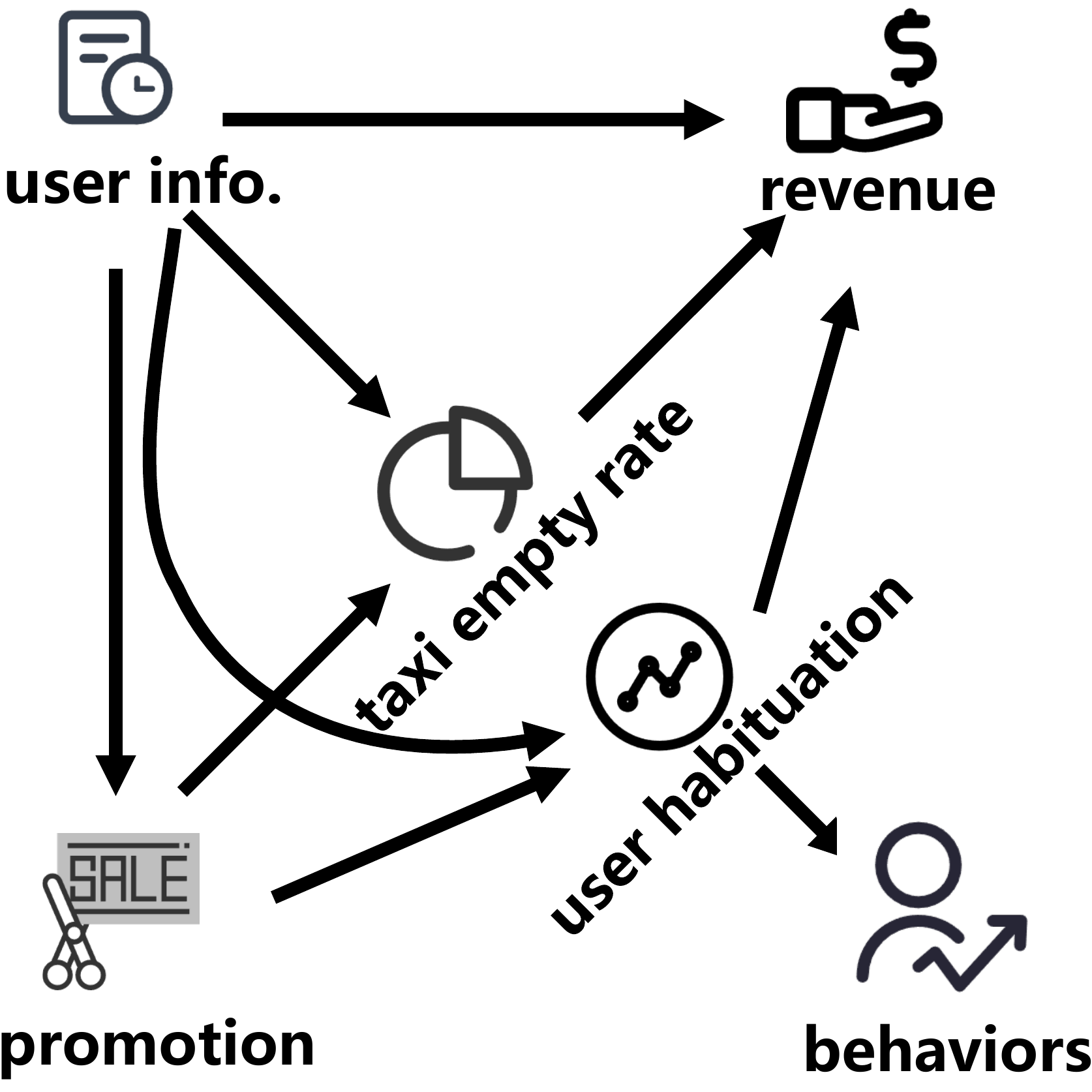}
}
\caption{Causal graphs in the long-term setting. The directed edges represent causal relationships. The gray nodes represent observed variables and the white nodes represent latent variables in which $x$ are the covariates, $t$ is the treatment, and $y$ is the long-term outcome. In Figure \ref{figure 1a}, $s$ is the valid surrogate. In Figure \ref{figure 1b}, $s_l$ is the latent surrogates, $s_o$ is the observed surrogates, and $p$ is the observed proxies of latent surrogates. Figure \ref{figure causal graph example} corresponds to Figure \ref{figure 1b} and serves as an example graph.}
\label{figure 1}
\end{center}
\end{figure}

In many real-world scenarios, however, without considering unobserved surrogates and by treating all short-term outcomes as surrogates,
existing methods estimate long-term effects in an idealistic and simplistic way like Figure \ref{figure 1a}.
As shown in Example \ref{eg: intor} and Figures \ref{figure 1b}, \ref{figure causal graph example}, in estimating the long-term effects of a promotion $t$ on the revenue $y$ on the ride-hailing platform, there exist unobserved surrogates $s_l$ (e.g., user habituation directly caused by the promotion) and lots of observed short-term variables, including the observed surrogates $s_o$ (e.g., regional empty taxi rate directly caused by the promotion) and \red{noisy} proxies $p$ of unobserved surrogates $s_l$ (e.g., users' behaviors \red{and users' cost of calling taxis} caused by user habituation).
The long-term revenue $y$ heavily relies only on the $s_o$ and $s_l$, but the $s_l$ can not be observed and the $s_o$ are indistinguishable from $p$ in multiple short-term outcomes. 
To work around this problem, existing works pose strong assumptions on the short-term outcomes $s_o$ and $p$ by excluding $s_l$ to obtain unbiased long-term outcome $y$, which is often impractical in real-world scenarios.

To tackle the challenges of unobserved surrogates and indistinguishable observed short-term outcomes ($s_o$ and $p$), we propose our method to identify long-term causal effects using LAtent SurrogatE Representation learning (LASER in short). Our method does not require accessing all valid surrogates nor distinguishing the observed surrogates from short-term outcomes. The generalized causal model is given in Figure \ref{figure 1b}.
In detail, we aim to learn the predictive surrogate representation based on an identifiable variational auto-encoder (iVAE). Given the learned representation, the long-term causal effects can be directly estimated by classic methods, e.g., the \red{Inverse Probability Weighting (}IPW\red{)} method. We also extend our work to two special cases - we observe all valid surrogates and we only observe proxies. In summary, our contributions are as follows:
\begin{itemize}
\item We propose our method LASER to overcome the challenge of estimating long-term causal effects in the case that the valid surrogates are partially or even completely unobserved but noisy proxies of latent surrogates can be observed.
\item Without the need to distinguish the observed surrogates and proxies of latent surrogates from short-term outcomes, we learn the surrogate representation in a unified way and provide the theoretically guaranteed identification of long-term causal effects.
\item We provide extensive experimental results on both real-world datasets and semi-synthetic datasets which demonstrate the effectiveness of our proposed method.
\end{itemize}

\section{Related Works}
\label{sec:related_works}

\textbf{Related to Traditional Causal Inference}: The traditional causal inference has been studied mainly in two languages: the graphical models \cite{pearl2009causal} and the potential outcome framework \cite{rubin1974estimating}. The most related method is the propensity score method in the potential outcome framework, e.g., IPW method \cite{IPWrosenbaum1983central, IPWrosenbaum1987model}. Moreover, several neural networks-based works \cite{johansson2016learning, shalit2017estimating, louizos2017causal, zhang2021treatment, shi2021invariant, assaad2021counterfactual} have been proposed, which mainly aim to learn a balanced representation or recover disentangled latent confounders. Different from these works above, we focus on the long-term setting and learn the latent surrogate representation instead of balanced representation or confounders.

\textbf{Related to Surrogates}: For many years, many works are interested in what is a valid surrogate that can reliably predict the long-term causal effects on treatment. Prentice criteria \cite{prentice1989surrogate} is the first criteria proposing statistical surrogate condition that only requires the treatment and long-term outcome are independent conditioning on the surrogate. Based on the principal stratification framework, principal criteria \cite{frangakis2002principal} additionally consider the property of causal necessity, i.e., the absence of the causal effects on the surrogate indicates the absence of causal effects on the long-term outcome. Based on the language of graphical models, strong surrogate criteria \cite{lauritzen2004discussion} argue that a surrogate is valid for the effect of a treatment on the long-term outcome if and only if treatment is an instrument for the effects of the surrogates on long-term outcome. Causal effect predictiveness (CEP) \cite{gilbert2008evaluating} was introduced to evaluate a principal surrogate. However, the surrogate paradox still exists when these criteria hold, which means the treatment has a positive effect on surrogates but a negative effect on the long-term outcome. Consistent surrogate and several variants \cite{chen2007criteria, ju2010criteria, yin2020novel} propose various conditions for testing a valid surrogate to avoid surrogate paradox.

Moreover, many works have been proposed to estimate long-term causal effects based on surrogates. SInd \cite{athey2019surrogate} combines multiple surrogates to robustly estimate long-term causal effects. \red{Cheng et~al. \cite{cheng2021long} propose the Long-Term Effect Estimation method (}LTEE\red{)} \red{by adapting} an RNN with double heads to learn the time-dependent link between time-dependent surrogates and the long-term outcome. \red{ Kallus et~al. \cite{kallus2020role} propose Efficient Estimation for Treatment Effects  (}EETE\red{) and}  analyze the benefit of surrogates in estimating causal effects and applies a nuisance estimator to long-term causal effects estimation. Some works \cite{athey2020combining, imbens2022long} also focus on the latent confounder problem in the long-term setting and propose causal effects estimators without the unconfoundedness assumption. Different from these works above,  we consider the observed short-term outcomes including observed surrogates and the proxies of latent surrogates, and in this case, these surrogate criteria do not hold, including traditional surrogacy assumption.

\textbf{Related to Variational Auto-encoders}: Our work is also related to the variational auto-encoders (VAE) \cite{vae2014original, rezende2014stochastic}. The VAE attempts to approximate an intractable posterior inference in the presence of latent variables. Its variants include $\beta$-VAE \cite{higgins2016beta}, \red{Hilbert-Schmidt information criterion-constrained VAE} \cite{lopez2018information}, $\sigma$-VAE \cite{pmlr-v139-rybkin21a}, \red{coupled conditional VAE} \cite{WANG2023} and so on. However, these variants do not guarantee the identifiability of latent variables. A recent variant, called identifiable VAE (iVAE) \cite{khemakhem2020variational}, approximates the posterior with additional supervised information to ensure recovery of the true latent variables.
Recently, Yao et.al \cite{yao2021learning} borrow the idea of iVAE and propose the \red{latent temporally causal processes estimation method} for modeling causal latent processes. 
In this paper, our method absorbs the core of iVAE, which not only learns the latent surrogate representation to guarantee the unbiased causal effects estimation but also provides the identification.

\section{Problem Definition}
\label{sec:pro_def}

Let $x \in \mathbb{R}^d$ be the covariates (also known as pre-treatment features or confounders). Let $t \in \{0,1\}$ denote a binary treatment, where $t=1$ indicates a unit receives the treatment (treated) and $t=0$ indicates a unit receives no treatment (control). Let $s_l  \in \mathbb{R}^a $ denote latent surrogates and $s_o  \in \mathbb{R}^b $ denote the observed surrogates, with the entire set of valid surrogates represented by concatenating $s=[s_o, s_l] \in \mathbb{R}^{n}$ where $n=a+b$. Let $p \in \mathbb{R}^c$ denote the proxies of latent surrogates, and for simplicity, we also denote the observed short-term outcomes as the mixture of observed surrogates and proxies of latent surrogates $m=[ s_o,p] \in \mathbb{R}^{b+c}$. Let $y \in \mathbb{R}$ be the long-term outcome. Let $y(1)$ and $y(0)$ denote the potential outcome under the treatment group and control group, respectively. Note that we can only observe one of the potential outcomes, and then $y = t*y(1) + (1-t)*y(0)$.

In this paper, given a set of historical data (observational data) $D_{obs} = \{ \langle x,t,m,y \rangle \}$, our goal is to estimate the average treatment effects (ATE) on the target population, i.e., experimental data, $D_{exp} = \{\langle x,t,m \rangle \}$ as follows:
\begin{equation}
    ATE = \tau_{exp} = \mathbb{E}[y|do(t=1)] - \mathbb{E}[y|do(t=0)],
\end{equation}
where $do(\cdot)$ means do-operation \cite{pearl2009causality}.

In this work, our assumptions include \red{Stable Unit Treatment Value Assumption (}SUTVA\red{)}, Overlap, and Unconfoundedness assumptions which are widely adopted in existing casual inference methods \cite{IPWrosenbaum1983central, athey2019surrogate, cheng2021long, kallus2020role, shi2021invariant}. We further assume the following assumptions hold:

 \begin{assumption} \label{ass:comparability}
     (Comparability Assumption \cite{athey2019surrogate}). Given the covariates $x$ and the valid surrogates $s_l, s_o$, the conditional distribution of $y$ in observational data is identical to that in experimental data. Formally, 
 \begin{equation*}
     p_{obs}(y|x,s_o, s_l) = p_{exp}(y|x,s_o, s_l),
 \end{equation*}
 where $p_{obs}$ and $p_{exp}$ denote the distribution on observational and experimental data, respectively.
 \end{assumption}

\begin{assumption} \label{ass:surrogacy}
    (Partially Latent Surrogacy Assumption). Given the covariates and the valid surrogates, including the observed surrogates $s_o$ and the latent surrogates $s_l$ of which we only have the proxies $p$, the treatment $t$ is independent of the long-term outcome $y$. Formally, 
\begin{equation*}
    y \perp\!\!\!\perp t | s_o, s_l, x.
\end{equation*}
\end{assumption}

Assumption \ref{ass:comparability} places the connection between observational and experimental data on the conditional distribution of $y$. It allows inferring $y$ in experimental data by making use of $y$ in observational data.

Assumption \ref{ass:surrogacy} is the core assumption in our paper, which implies that given covariates the \textit{valid surrogates} contain all causal information between the treatment and the long-term outcome. This assumption does not require full observation of the valid surrogates and even does not require observing one of the surrogates as long as the proxies are observed. Please note that this assumption is much weaker than the original surrogacy assumption in \cite{athey2019surrogate}, i.e., $y \perp\!\!\!\perp t |s,x$, which requires full observation of surrogates to totally block the causal path between the treatment and the long-term outcome.

It is challenging to estimate long-term causal effects when surrogates are partially observed and mixed with noisy proxies, which, however, is a very common scenario in reality (see Example \ref{eg: intor}). Surprisingly, we find that noisy proxies of latent surrogates can still be useful in estimating long-term causal effects. Thus, we develop a representation learning method to jointly process noisy proxies and partially observed surrogates (see Section \ref{sec:method}) and provide theoretically guaranteed and unbiased estimation (see Section \ref{sec:the_ana}).

\section{Methodology}
\label{sec:method}

\begin{figure*}[!ht]
\begin{center}
\subfigure[Inference network, $q(s|x,m,t)$.]{ \label{figure 2a}
    \includegraphics[width=0.45\textwidth]{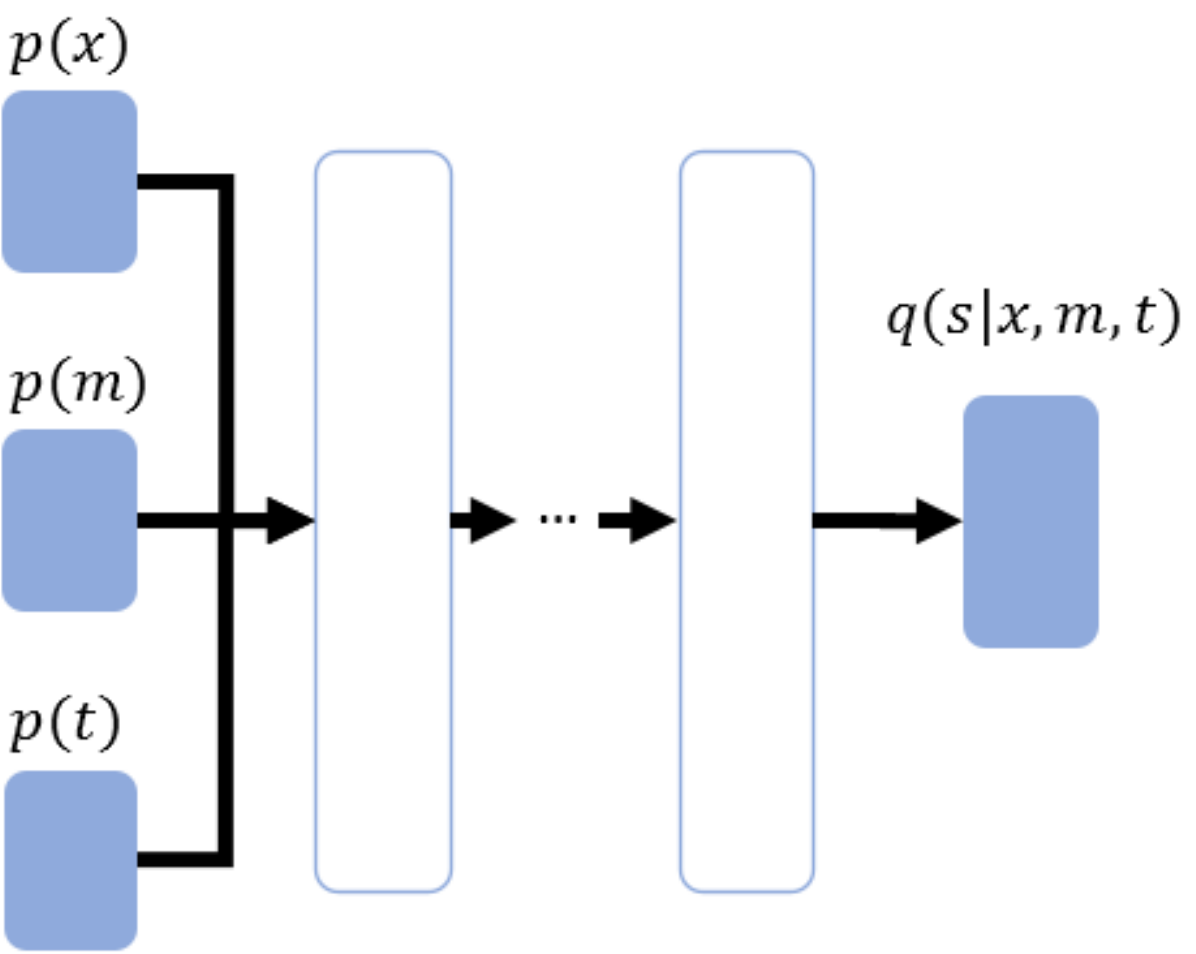}
}
\hspace{0.02\textwidth}
\subfigure[Generative network, $p(m|s)$ and $p(y|s)$.]{   \label{figure 2b}
    \includegraphics[width=0.45\textwidth]{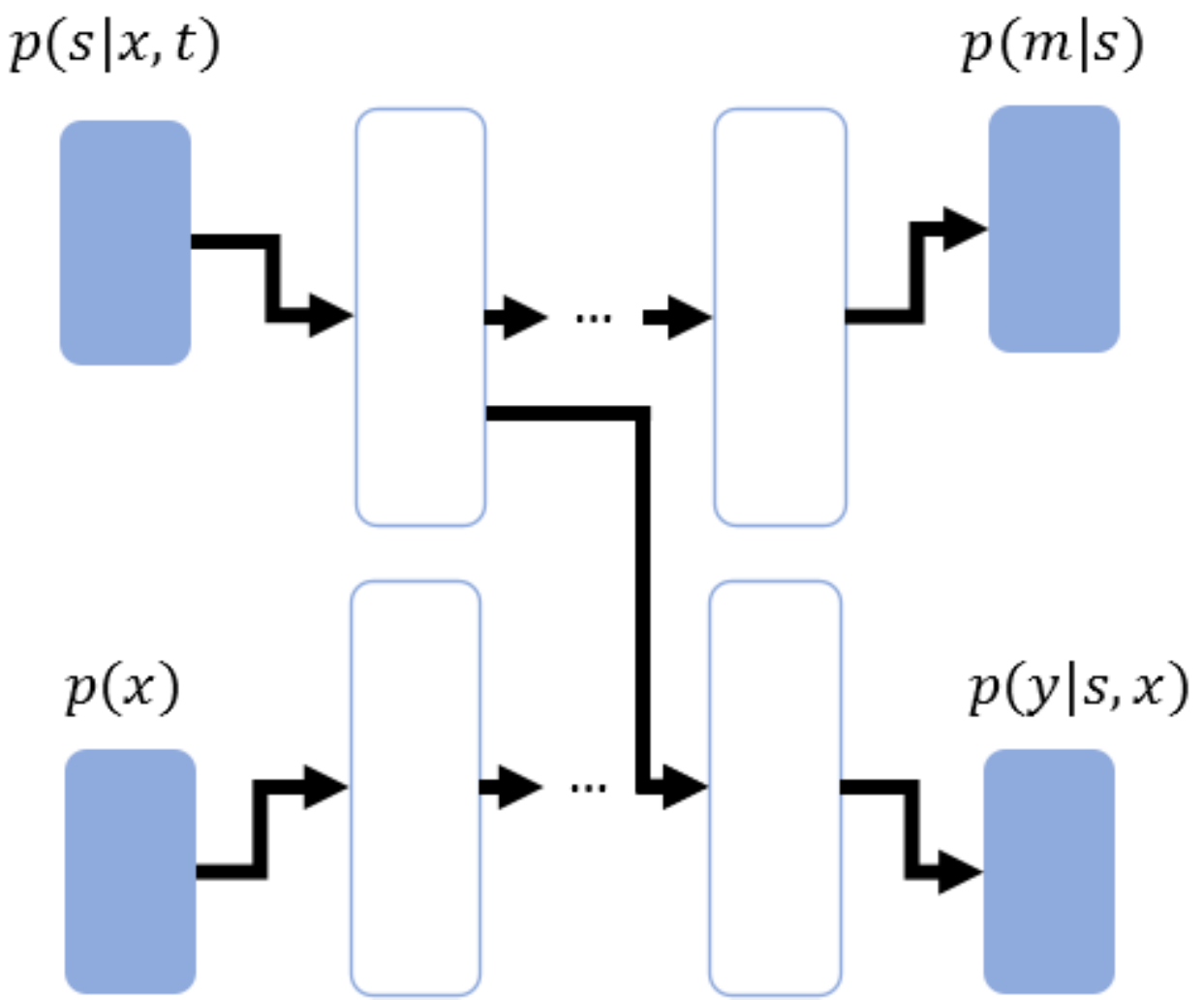}
}
\caption{Overall architecture of inference network and generative network. White nodes correspond to parametric deterministic neural network transitions, and blue nodes correspond to samples drawn from respective distributions.}
\label{figure 2}
\end{center}
\end{figure*}

In this section, we aim to devise a practical LAtent SurrogatEs Representation learning method (LASER in short) to construct a predictive surrogate representation and further estimate the long-term causal effects, jointly taking the mixture of observed surrogates and proxies of latent surrogates as inputs. LASER generally consists of the following two steps: the iVAE-based representation learning step and the IPW-based causal effects estimation step.

In the \emph{representation learning} step, we employ iVAE to construct a predictive surrogate representation from the mixture of observed surrogates and the proxies of latent surrogates as inputs. Specifically, considering the causal graph in Figure \ref{figure 1b}, we reconstruct $s$ from $s_o \cup p$, by incorporating $x$ and $t$ as additional supervised information to approximate the conditional prior $p(s|x,t)$. Such additional supervised information not only improves the quality of the learned representations but also guarantees the identifiability of the model.

We further devise an end-to-end neural network architecture for the representation learning as shown in Figure \ref{figure 2}. The design of inference and generative network follows the distribution factorization as shown in Figure \ref{figure 1b}. For the inference network, the covariates $x$ and the treatment $t$ as additional supervised information are input together with $m$ to approximate appropriate prior $p(s|x,t)$, which importantly guarantees the power of approximation. For the generative network, besides the approximation of posterior $p(m|s)$, we also employ an \red{ MultiLayer Perceptron (}MLP\red{)} to approximate $p(y|s,x)$. Considering the partially latent surrogacy assumption \ref{ass:surrogacy}, training such an MLP at the same time will not only encourage the learned $s$ approximating the valid surrogates but also avoid the extra regression step of fitting $\mathbb{E}[y|s,x]$ after learning the surrogates $s$.

Following existing VAE-based works, we choose the prior $p(s|x, t)$ as standard Gaussian distribution, which can be extended to exponential family distributions as discussed in Section \ref{sec:the_ana}, and thus each of the dimensions of $s$ has two sufficient statistics, the mean and variance:
\begin{equation}
  \begin{aligned}
    p(s|x,t) = \prod_{i=0}^{n-1} \mathcal{N}(s_{i}|0,1),
  \end{aligned}
\end{equation}
where $n$ is the parameter that determines the dimensions of surrogates $s$.

In the inference network, the variational approximation of the posterior is defined as:
\begin{equation}
  \begin{aligned}
    q(s|m,x,t) = \prod_{i=0}^{n-1} \mathcal{N}(\mu=\hat{\mu}_{s_{i}},\sigma^2=\hat{\sigma}_{s_{i}}^2),
\end{aligned}
\end{equation}
where $\hat{\mu}_{s_{i}}$ and $\hat{\sigma}_{s_{i}}^2$ are the mean and variance of the Gaussian distribution parametrized by the inference network.

In the generative network, for a continuous outcome, we parametrize the probability distribution as a Gaussian distribution with its mean given by an MLP and a fixed variance $v^2$. For a discrete outcome, we use a Bernoulli distribution parametrized by an MLP similarly:
\begin{equation}
  \begin{aligned}
    p(m|s) = \mathcal{N}(\mu=\hat{\mu}_m,\sigma^2=v_m^2) 
    \hspace{4mm}& \text{or} \hspace{4mm}
    p(m|s) =  \mathbf{Bern}(\pi = \hat{\pi}_m), \\ 
    p(y|s,x) = \mathcal{N}(\mu=\hat{\mu}_y,\sigma^2=v_y^2) 
    \hspace{4mm} & \text{or} \hspace{4mm}
    p(y|s,x) =  \mathbf{Bern}(\pi = \hat{\pi}_y), 
\end{aligned}
\end{equation}
where for the continuous case $\hat{\mu}_m$ and $\hat{\mu}_y$ are the mean of the Gaussian distribution parametrized by the generative network similar to inference network, and $v_m^2$ and $v_y^2$ are the fixed variance of Gaussian distribution, and for the discrete case  $\hat{\pi}_m$ and $\hat{\pi}_y$ are the mean of Bernoulli distribution similarly parametrized by the network.

We can now form the objective function $\mathcal{L}$ for the inference and generative networks, consisting of the negative variational Evidence Lower BOund (ELBO) of the graphical model and the negative marginal log-likelihood of $y$ for reconstruction as follows:
\begin{equation}
  \begin{aligned}
    \mathcal{L}  =& -ELBO + \mathcal{L}_y \\
                 =& - \mathbb{E}_{q_{D_{obs} \cup D_{exp}}} [
                 \mathbb{E}_{q(s|m,x,t)} [ \log p(s|x,t)  
                + \log p(m|s) - \log q(s|m,x,t) ]] \\
                &- \mathbb{E}_{q_{D_{obs}}} [ \mathbb{E}_{q(s|m,x,t)} [  \log p(y|s,x) ]] ,
\end{aligned}
\end{equation}
where $q_{D_{obs} \cup D_{exp}}$ and $q_{D_{obs}}$ are the empirical data distributions given by dataset $D_{obs} \cup D_{exp}$ and $D_{obs}$ respectively. The ELBO derivation is given in \ref{appendix: elbo}.

In the \emph{causal effects estimation} step, we simply utilize the IPW method to estimate the long-term causal effects. Specifically, given experimental data $D_{exp}$, we estimate the long-term causal effects $\hat{\tau}_{exp}$ as follows: 
\begin{equation} \label{eq:estimated ce}
  \begin{aligned}
    \hat{\tau}_{exp} = \mathbb{E} \left[ \hat{y}  \frac{t}{e(x)} -\hat{y}  \frac{1-t}{1-e(x)}  \right],
\end{aligned}
\end{equation}
where $\hat{y}$ is the output of the neural networks with input $D_{exp}$ and $e(\cdot)$ denotes the propensity score $\mathbb{E}[t|x]$.

It is worth noting that LASER does not need to distinguish between observed surrogates and proxies of latent surrogates within the mixed short-term outcomes. Instead, we use all the observed short-term outcomes, including $s_o$ and $p$, to learn the complete surrogates $s_o$ and $s_l$. Our method provides a unified way to process both the surrogates and proxies of latent surrogates. In the next section, we will discuss the identifiability to guarantee the correctness of our model.

\section{Theoretical Analysis}
\label{sec:the_ana}

In this section, we present the identifiability of our model and long-term causal effects. Note that if we correctly identify the valid surrogates, the long-term causal effects can be identified based on the IPW method. 

For simplicity, we externally define the proxies $p_o$ of $s_o$ without noise as $p_o = s_o$. Thus the mixture $m$ can be viewed as $m=[s_o, p]=[p_o,p]$, which only consists of the proxies. In this way, both $s_o$ and $s_l$ can be considered as the latent variables, while both $p_o$ and $p$ can be observed.

Let $\theta = (\mathbf{T},\mathbf{f},\bm{\lambda})$ be the parameters on $\Theta$ of the following conditional generative models:
\begin{equation}\label{dgp}
 \begin{aligned}
    p(s,\varepsilon,m|x,t) 
    & = p_{\mathbf{T},\bm{\lambda}}(s_o, s_l, \varepsilon|x,t)p_{\mathbf{f}}(p_o, p|s_o, s_l,\varepsilon) \\
    & = p_{\mathbf{T},\bm{\lambda}}(s,\varepsilon|x,t)p_{\mathbf{f}}(m|s,\varepsilon),
 \end{aligned}
\end{equation}
where we define the following injective function $\mathbf f$:
\begin{equation} \label{phi definition}
    m = \mathbf f(s,\varepsilon)
\end{equation}
in which we assume that $\varepsilon \in \mathbb{R}^n$ is \red{a} conditional independent noise variable with probability density function $p_\varepsilon(\varepsilon)$, i.e., $\varepsilon_i$ is mutually independent given $(x,t)$. \red{The proxies $m$ are allowed to have different practical meaning from the surrogates $s$, but must be generated by $s$ and $m$ as valid proxies.}

To specify the correlation between latent variables $s,\varepsilon$ and observed \red{variables $x,t$}, we assume the conditional distribution $p_{\mathbf{T},\bm{\lambda}}(s, \varepsilon|x,t)$ is conditional factorial with an exponential family distribution:

\begin{assumption} \label{ass: exp dist}
    The correlation between  $s,\varepsilon$ and  $x,t$ is characterized by:
    \begin{equation} \label{exp dist definition}
      \begin{aligned}
        & p_{\mathbf{T},\bm{\lambda}}(s, \varepsilon|x,t) \\
        = &
        p_{\mathbf{T},\bm{\lambda}}(s|x,t)p_{\mathbf{T},\bm{\lambda}}(\varepsilon|x,t)
        \\
        = & (\prod_{i=1}^{n} \frac{Q_i(s_i)}{Z_i(x,t)}
        \exp [\sum_{j=1}^{k} T_{ij}(s_i)\lambda_{ij}(x,t)])
         \times (\prod_{i=n+1}^{2n} \frac{Q_i(\varepsilon_i)}{Z_i(x,t)}
        \exp [\sum_{j=1}^{k} T_{ij}(\varepsilon_i)\lambda_{ij}(x,t)]),
      \end{aligned}
    \end{equation}
    where $Q_i$ is the base measure, $Z_i(x,t)$ is the normalizing constant, $\mathbf{T_i}=(T_{i,1},...,T_{i,k})$ are sufficient statistics, and $\mathbf{\lambda}_i(x,t)=(\lambda_{i,1}(x,t),...,\lambda_{i,k}(x,t))$ are the $x,t$ dependent parameters, and $k$ is the dimension of each sufficient statistic.
\end{assumption}

\red{Assumption \ref{ass: exp dist} specifies that the conditioning on $(x,t)$ is through an arbitrary function $\bm{\lambda}(x,t)$, which is crucial to make the latent surrogates identifiable. Note that the exponential families include many common distributions such as normal, exponential, gamma, beta, and Poisson, so Assumption \ref{ass: exp dist} is not very restrictive.} Such an assumption is widely used in VAE-based methods such as \cite{louizos2017causal, zhang2021treatment, khemakhem2020variational}, and many of them usually further specify that the (conditional) prior is Gaussian. In our implementation, we also use Gaussian distribution as \red{the conditional} prior.

Then, we consider the following equivalence relation on the set of parameters $\Theta$ from \cite{khemakhem2020variational}:

\begin{definition} Let $\sim_A$ be the binary relation on $\Theta$ defined as follows: 
\begin{equation} 
    \begin{aligned}
        & (\mathbf{f},\mathbf{T},\bm{\lambda})
        \sim_A (\mathbf{\hat f}, \mathbf{\hat T},\bm{\hat \lambda}) 
         \Leftrightarrow \\
        & \exists A, c \hspace{1mm} | \hspace{1mm} \mathbf{T}(\mathbf{f}^{-1}(x))= A \mathbf{\hat T}^(\mathbf{\hat f}^{-1}(x))+ \mathbf{c}, \forall x \in \mathcal{X} 
    \end{aligned}
\end{equation}
    where $A$ is a invertible matrix and $ \mathbf{c}$ is a vector, and $\mathcal{X}$ is the domain of $x$.
\end{definition}

When the underlying model parameters $\theta=(\mathbf{f},\mathbf{T},\bm{\lambda})$ can be recovered by perfectly fitting the conditional distribution $p_{\theta}(m|x,t)$, the joint distribution $p_{\theta}(s,\varepsilon,m|x,t)$ also can be recovered. This further implies the recovery of conditional prior $p_{\mathbf{\hat T},\bm{\hat \lambda}}(s,\varepsilon|x,t)$ and the recovery of latent surrogates $\hat{s}$ (and noises $\hat{\varepsilon}$) up to some simple transformations. We conclude the identifiability of our model as follows:

\begin{theorem}\label{theorem:recover latent}
Assume that the data we observed are generated following Eq. \ref{dgp} and \ref{phi definition} and suppose Assumption \ref{ass: exp dist} holds, and suppose the following conditions hold: (1) The sufficient statistics $T_{i,j}$ are differentiable almost everywhere, and $(T_{i, j})_{1\leq j \leq  k}$ are linearly independent on any subset of the domain of $(x,t)$ of measure greater than zero. (2) There exist $2nk+1$ distinct points $(x_0,t_0), ... , (x_{2nk},t_{2nk})$ such that the matrix
    \begin{equation*}
    \small
         L =  (\mathbf{\lambda}((x_1,t_1))- \mathbf{\lambda}((x_0,t_0)),...,
         \mathbf{\lambda}((x_{2n},t_{2n}))-\mathbf{\lambda}((x_0,t_0)))
    \end{equation*}
    of size $2nk \times 2nk$ is invertible.
Then the parameters $(\mathbf{T},\mathbf{f},\mathbf{\lambda})$ are $\sim_A -identifiable$.
\end{theorem}

The proof is in \ref{appendix: proof of latent identifiability}. Note that the last assumption in Theorem \ref{theorem:recover latent} requires there \red{exist} $2nk+1$ distinct points $(x_i,t_i)$. \red{The assumption} is reasonable \red{for two reasons.} First, based on the Overlap assumption\red{, i.e., $0<p(t|x)<1, \forall t \in \{0,1\}$, we only require that there exists $nk+1$ distinct points $x_i$. Further, in many scenarios the space $\mathcal{X}$ of multivariate covariate $x$ is large enough so that we can easily collect sufficient enough different covariates.}

Theorem \ref{theorem:recover latent} indicates that under mild assumptions, we can recover the $\hat{s}$ and $\hat{\varepsilon}$ which are equal to a permutation (the matrix $A$) and a point-wise nonlinearity (in the form of $\mathbf{T}$ and $\mathbf{ \hat T}$) of the original latent surrogates $s$ and noises $\varepsilon$. Based on such recovered $\hat s$, we can further identify the long-term causal effects shown in the following theorem.

\begin{theorem}\label{theorem:ce}
Suppose assumptions SUTVA, Overlap, Unconfoundedness, assumption \ref{ass:comparability}, assumption \ref{ass:surrogacy}, and Theorem \ref{theorem:recover latent} hold, 
then the causal effect is equal to the estimated causal effect in Eq. \ref{eq:estimated ce}: $\hat{\tau}_{exp} = \tau_{exp}$.
\end{theorem}

The proof is in \ref{appendix: proof of ce identifiability}. Theorem \ref{theorem:ce} guarantees that given enough information that makes $s$ identifiable, the estimated ATE can also be identified under some mild assumptions which are usually assumed in causal inference. It theoretically guarantees the correctness of our model, providing a feasible technology of long-term causal effects estimation via learning latent surrogates. We also extend Theorem \ref{theorem:ce} to the following two special cases:

\begin{figure}[!t]
\begin{center}
\subfigure[Special case 1 (same as Figure \ref{figure 1a}).]{ \label{figure 3a}
    \includegraphics[width=0.35\textwidth]{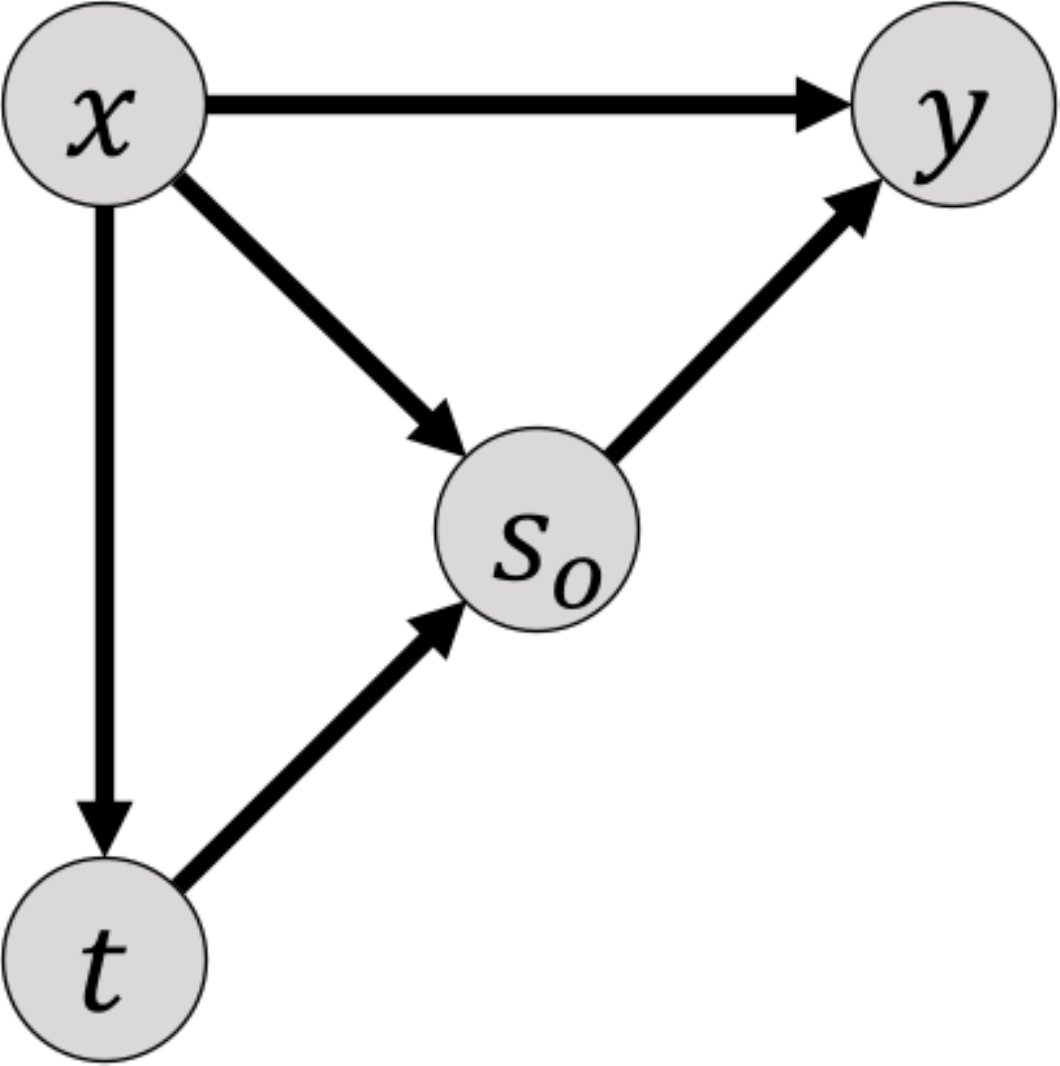}
}
\hspace{10mm}
\subfigure[Special case 2.]{ \label{figure 3b}
    \includegraphics[width=0.35\textwidth]{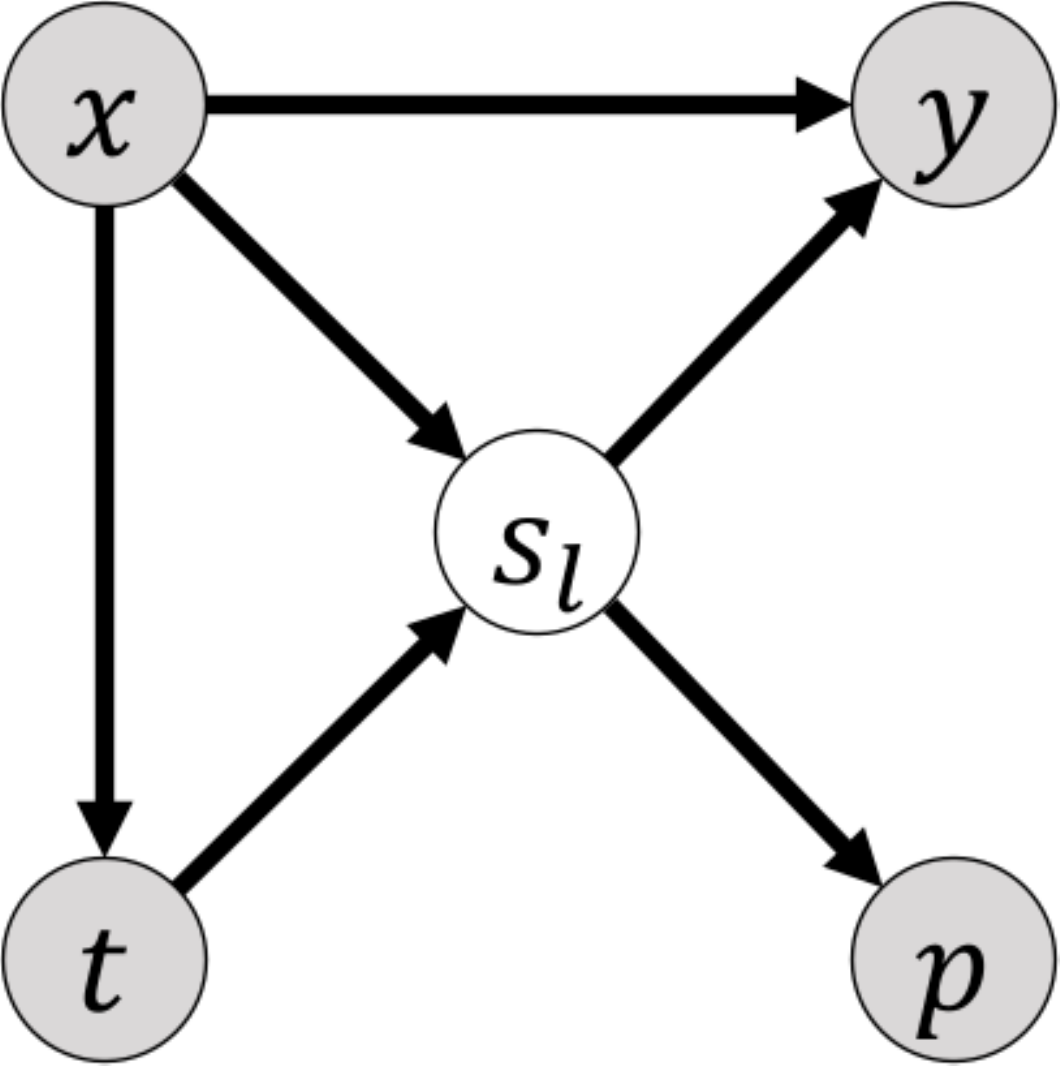}
}
\vskip -0.1in
\caption{Two special cases of our causal graph. The gray nodes represent observed variables and the white nodes represent latent variables. In the case of Figure \ref{figure 3a}, we can observe the whole valid surrogates. In the case of Figure  \ref{figure 3b}, we only observe the proxies of surrogates.}
\label{figure 3}
\end{center}
\vskip -0.2in
\end{figure}

\begin{corollary} \label{corollary:special case}
Suppose Theorem \ref{theorem:ce} holds and the data generation progress follows the causal graph in Figure \ref{figure 3a} or \ref{figure 3b}, then the causal effect is also equal to the estimated causal effect.
\end{corollary}

The proof is in \ref{app: proof of special cases}.  \red{Note that Figures \ref{figure 3a} and \ref{figure 3b} are only special cases under which our method LASER works well. In Figure \ref{figure 3a}, the proxies can be considered completely the same as observed surrogates, i.e., $s_o = p$. In Figure \ref{figure 3b}, there are no observed surrogates and all observed short-term outcomes are proxies.}

Corollary \ref{corollary:special case} indicates that regardless we observe the valid surrogates, the proxies of latent surrogates, or the mixed case, our method is able to estimate the long-term causal effects robustly, which guarantees practicability and flexibility of our representation learning based estimation method. In many real-world scenarios, we have a myriad of observed short-term variables, making it extremely difficult to distinguish between surrogates and proxies. Theorem \ref{theorem:ce} and Corollary \ref{corollary:special case} allow inferring long-term causal effects without considering the complex causal structure among short-term variables and identifying which are surrogates or proxies.

It is worth noting that the traditionally assumed causal graph like Figure \ref{figure 3a} is also a special case of our model. It means that without testing whether the standard surrogacy assumption $y \perp\!\!\!\perp t | s,x$ holds, our model can be an out-of-box tool to estimate long-term causal effects for many real-world applications.

\section{Experiments}
\label{sec:exp}

 In this section, we validate the proposed \textbf{LASER} on real-world and semi-synthetic datasets. In detail, we verify the effectiveness of our method on the real-world dataset \red{in Section \ref{sbsection: real-world}} and we further evaluate the correctness of the theorem analysis with the help of the semi-synthetic datasets \red{in Section \ref{sbsection: semi-synthetic} and \ref{sbsection: semi-synthetic 2}}. The real-world dataset is generously provided by the Chinese largest mobility technology platform and the semi-synthetic datasets are simulated based on \red{the} Infant Health and Development Program (IHDP) \cite{hill2011bayesian}.
 
 \textbf{Baseline Methods:} We consider several baseline methods including the state-of-the-art model based on neural networks like LTEE \cite{cheng2021long} and some statistical models like SInd \cite{athey2019surrogate} and Back-Door adjustment method (BD) \cite{pearl1993bayesian}. Some baseline methods are proposed to deal with different settings, and we will clarify them as follows:
 \begin{itemize}
\item LTEE \cite{cheng2021long}: LTEE uses an RNN with double heads to capture the relationship of time-dependent surrogates respectively in the treatment and control group, and then estimate CATE by the predicted long-term outcome. We use the code available at \url{https://github.com/GitHubLuCheng/LTEE}. It requires $x,t,m,y$ in observational data and $x$ in experimental data.
\item SInd-Linear and SInd-MLP \cite{athey2019surrogate}: SInd uses a predictor named "Surrogate Index" to predict the long-term outcome using surrogates, and the estimate CATE by the IPW method. We use linear regression (SInd-Linear) or an MLP (SInd-MLP) as the predictor to capture the relationship between surrogates and the long-term outcome. It requires $x,m,y$ in observational data and $x,t,m$ in experimental data.
\item \red{BD-Linear and BD-MLP \cite{pearl1993bayesian}: We employ the back-door adjustment method using the fitted experimental long-term outcome by SInd. Specifically, we obtain the fitted long-term outcome $\hat{y}$ of experimental data by SInd and perform two regressions of $\hat y$ on the back-door adjustment set $x$, i.e., $\mathbb E[y|t=1,x]$ and $\mathbb E[y|t=1,x]$, using linear regressions (BD-Linear) or MLPs (BD-MLP). It requires $x,m,y$ in observational data and $x,t,m$ in experimental data.}
\item EETE \cite{kallus2020role}: EETE proposes an estimator and inferential method based on flexible machine learning methods to estimate nuisance parameters that appear in the influence functions. We implement its estimator in Python. To estimate causal effects, it requires $x,t,m,y$ in observational data and $x,t,m$ in experimental data.
\end{itemize}

 \textbf{Implementation details:} We implement the basic version denoted by \textbf{LASER} and a modified version denoted by \textbf{LASER-GRU}. In LASER-GRU, we modify the extra MLP in the generative model to capture the time-dependent relationship between the long-term outcome and the specific surrogates $s_{o^{'}}$ that has the same practical meaning (e.g., the first $14$ days' revenues when $y$ is the $90$th day's revenue). We apply a GRU by inputting $s_{o^{'}}$ to generate the time-dependent representation $r$. Thus the modified extra MLP approximates: $p(y|s_l,x,r)$ where $r=GRU(s_{o^{'}})$.
 Our implementation is based on the code of iVAE. We choose the Leaky-ReLu as the activation function in our model and Adam is adopted to solve the optimization problem. All the experiments can be run on a single 8GB GPU of NVIDIA Tesla P40. Our code will be available after this paper is accepted.

 \textbf{Metric:} We report the mean and the standard deviation of mean absolute percentage error (MAPE) of ATE on experimental data in $n$-time runnings, where $MAPE=  | \frac{\tau_{exp} - \hat{\tau}_{exp}}{\tau_{exp}}  |$. The values presented are averaged over at least 5 replicated with different random seeds. The standard deviation is in the subscript.

\subsection{Real-World Dataset}
\label{sbsection: real-world}

\begin{table}[!ht] 
\caption{Detailed Information of Real-world Dataset}\label{tb:detailed real dataset}
\begin{center}
\begin{tabular}{c|cl|c}
\hline
Subject & \red{Property} &: \red{ Number}   & \red{Exist In}     \\ 
\hline
 $x$ & \red{dimension} &: 15  & \red{$D_{obs}, D_{exp}$ }     \\ 
\hline
 $m$ (all surrogates) & \red{dimension} &: 105  & \red{$D_{obs}, D_{exp}$ }      \\ 
\hline
 $m$ (revenues only) & \red{dimension} &: 14  & \red{$D_{obs}, D_{exp}$ }      \\ 
\hline
\red{ $y$}  & \red{dimension} &: \red{1}  & \red{ $D_{obs}$ }      \\ 
\hline
observational data (per strategy)
& \red{sample size} &: 4592  &\red{ $D_{obs}$ }      \\ 
\hline
experimental data (per strategy) 
& \red{sample size} &: 1008 &  \red{ $D_{exp}$ }       \\ 
\hline
\end{tabular}
\end{center}
\vskip -0.2in
\end{table}

\begin{table}[!ht]
\renewcommand\arraystretch{1.05}
\caption{MAPE on Real-world dataset (Revenues  Only)}
\vskip -0.1in
\label{tb:real-world result Benefits Only}
\begin{center}
\scalebox{0.5}{
\begin{tabular}{cccccccccc} 
	& \ & LASER & LASER-GRU & SInd-Linear   & SInd-MLP   & LTEE  & EETE & \red{BD-Linear} & \red{BD-MLP}     \\ \hline
	Strategy 
	 & City& mean             $_{\pm \text{ std}}$ & mean   $_{\pm \text{ std}}$  & mean   $_{\pm \text{ std}}$    & mean    $_{\pm \text{ std}}$   & mean $_{\pm \text{ std}}$  & mean $_{\pm \text{ std}}$ & \red{mean $_{\pm \text{ std}}$ } & \red{ mean $_{\pm \text{ std}}$ }\\ \hline
	\multirow{5}{*}{Strategy1} 
	 & All & 0.0179          $_{\pm \text{ 0.0095}}$ & 0.0219 $_{\pm \text{ 0.0095}}$ & 0.2070 $_{\pm \text{ 0.0095}}$ & 0.1253 $_{\pm \text{ 0.0157}}$ & \textbf{0.0151} $_{\pm \text{ 0.0142}}$ & 0.0265 $_{\pm \text{ 0.0149}}$ & 0.2086 $_{\pm \text{ 0.0092}}$ & 0.1808 $_{\pm \text{ 0.0324}}$  \\ 
	 & A   & \textbf{0.0306} $_{\pm \text{ 0.0195}}$ & 0.0328 $_{\pm \text{ 0.0129}}$  & 0.2406 $_{\pm \text{ 0.0277}}$ & 0.1170 $_{\pm \text{ 0.0583}}$ & 0.0562          $_{\pm \text{ 0.0251}}$ & 0.0682 $_{\pm \text{ 0.0199}}$ & 0.2246 $_{\pm \text{ 0.0262}}$ & 0.1817 $_{\pm \text{ 0.0259}}$ \\ 
	 & B   & \textbf{0.0469} $_{\pm \text{ 0.0286}}$ & 0.0563 $_{\pm \text{ 0.0151}}$ & 0.1868 $_{\pm \text{ 0.0164}}$ & 0.1855 $_{\pm \text{ 0.0328}}$ & 0.1069          $_{\pm \text{ 0.0518}}$ & 0.2112 $_{\pm \text{ 0.0744}}$ & 0.2120 $_{\pm \text{ 0.0294}}$ & 0.2060 $_{\pm \text{ 0.0571}}$ \\ 
	 & C   & \textbf{0.0232} $_{\pm \text{ 0.0140}}$ & 0.0322 $_{\pm \text{ 0.0138}}$  & 0.1481 $_{\pm \text{ 0.0263}}$ & 0.1146 $_{\pm \text{ 0.0285}}$ & 0.0839          $_{\pm \text{ 0.0454}}$ & 0.0311 $_{\pm \text{ 0.0295}}$  & 0.1961 $_{\pm \text{ 0.0398}}$ & 0.2056 $_{\pm \text{ 0.0544}}$ \\ 
	 & D   & \textbf{0.0337} $_{\pm \text{ 0.0229}}$ & 0.0373 $_{\pm \text{ 0.0208}}$  & 0.1191 $_{\pm \text{ 0.0147}}$ & 0.0926 $_{\pm \text{ 0.0285}}$ & 0.0524          $_{\pm \text{ 0.0437}}$ & 0.0923 $_{\pm \text{ 0.0084}}$ & 0.1808 $_{\pm \text{ 0.0474}}$ & 0.1923 $_{\pm \text{ 0.0583}}$ \\ \hline
	\multirow{5}{*}{Strategy2} 
	 & All & 0.0283 $_{\pm \text{ 0.0235}}$ & \textbf{0.0167} $_{\pm \text{ 0.0119}}$  & 0.2591 $_{\pm \text{ 0.0141}}$ & 0.1584 $_{\pm \text{ 0.0184}}$ & 0.0334 $_{\pm \text{ 0.0191}}$ & 0.0636          $_{\pm \text{ 0.0188}}$  & 0.1942 $_{\pm \text{ 0.0529}}$ & 0.2346 $_{\pm \text{ 0.0095}}$  \\ 
	 & A   & \textbf{0.0127} $_{\pm \text{ 0.0043}}$ & 0.0301 $_{\pm \text{ 0.0188}}$  & 0.2569 $_{\pm \text{ 0.0302}}$ & 0.1300 $_{\pm \text{ 0.0136}}$ & 0.0649 $_{\pm \text{ 0.0515}}$ & 0.1091          $_{\pm \text{ 0.0287}}$ & 0.2032 $_{\pm \text{ 0.0548}}$ & 0.2470 $_{\pm \text{ 0.0332}}$  \\ 
	 & B   & 0.0686          $_{\pm \text{ 0.0430}}$ & 0.0565 $_{\pm \text{ 0.0468}}$  & 0.2160 $_{\pm \text{ 0.0468}}$ & 0.2160 $_{\pm \text{ 0.0842}}$ & 0.1447 $_{\pm \text{ 0.0279}}$ & \textbf{0.0508} $_{\pm \text{ 0.0496}}$  & 0.2048 $_{\pm \text{ 0.0541}}$ & 0.2392 $_{\pm \text{ 0.0557}}$  \\ 
	 & C   & \textbf{0.0192} $_{\pm \text{ 0.0100}}$ & 0.0412 $_{\pm \text{ 0.0241}}$  & 0.2092 $_{\pm \text{ 0.0413}}$ & 0.1753 $_{\pm \text{ 0.0543}}$ & 0.0899 $_{\pm \text{ 0.0568}}$ & 0.0234          $_{\pm \text{ 0.0194}}$  & 0.2053 $_{\pm \text{ 0.0528}}$ & 0.2428 $_{\pm \text{ 0.0542}}$ \\ 
	 & D   & 0.0420 $_{\pm \text{ 0.0333}}$ & \textbf{0.0229} $_{\pm \text{ 0.0119}}$  & 0.1906 $_{\pm \text{ 0.0287}}$ & 0.1293 $_{\pm \text{ 0.0484}}$ & 0.0645 $_{\pm \text{ 0.0835}}$ & 0.0682          $_{\pm \text{ 0.0337}}$   & 0.2038 $_{\pm \text{ 0.0511}}$ & 0.2457 $_{\pm \text{ 0.0541}}$  \\ \hline
	\multirow{5}{*}{Strategy3} 
	 & All & 0.0471 $_{\pm \text{ 0.0353}}$ & \textbf{0.0393} $_{\pm \text{ 0.0213}}$  & 0.3301 $_{\pm \text{ 0.0094}}$ & 0.2011 $_{\pm \text{ 0.0327}}$ & 0.0476 $_{\pm \text{ 0.0246}}$ & 0.1774 $_{\pm \text{ 0.0373}}$  & 0.2170 $_{\pm \text{ 0.0641}}$ & 0.2893 $_{\pm \text{ 0.0502}}$  \\ 
	 & A   & \textbf{0.0384} $_{\pm \text{ 0.0215}}$ & 0.0504 $_{\pm \text{ 0.0491}}$  & 0.3595 $_{\pm \text{ 0.0305}}$ & 0.1374 $_{\pm \text{ 0.0440}}$ & 0.0601 $_{\pm \text{ 0.0381}}$ & 0.2903 $_{\pm \text{ 0.0532}}$   & 0.2289 $_{\pm \text{ 0.0735}}$ & 0.2917 $_{\pm \text{ 0.0592}}$  \\ 
	 & B   & 0.0833 $_{\pm \text{ 0.0403}}$ & \textbf{0.0773} $_{\pm \text{ 0.0371}}$  & 0.2714 $_{\pm \text{ 0.0176}}$ & 0.2754 $_{\pm \text{ 0.0818}}$ & 0.1034 $_{\pm \text{ 0.0376}}$ & 0.3425 $_{\pm \text{ 0.0885}}$  & 0.2321 $_{\pm \text{ 0.0717}}$ & 0.3101 $_{\pm \text{ 0.0603}}$\\ 
	 & C   & \textbf{0.0583} $_{\pm \text{ 0.0426}}$ & 0.0681 $_{\pm \text{ 0.0253}}$  & 0.2600 $_{\pm \text{ 0.0322}}$ & 0.1839 $_{\pm \text{ 0.0499}}$ & 0.1251 $_{\pm \text{ 0.0617}}$ & 0.1743 $_{\pm \text{ 0.0644}}$  & 0.2341 $_{\pm \text{ 0.0700}}$ & 0.3009 $_{\pm \text{ 0.0585}}$   \\ 
	 & D   & 0.0788 $_{\pm \text{ 0.0279}}$ & \textbf{0.0638} $_{\pm \text{ 0.0420}}$  & 0.2087 $_{\pm \text{ 0.0440}}$ & 0.1480 $_{\pm \text{ 0.0823}}$ & 0.1414 $_{\pm \text{ 0.0294}}$ & 0.1130 $_{\pm \text{ 0.0359}}$  & 0.2325 $_{\pm \text{ 0.0688}}$ & 0.2881 $_{\pm \text{ 0.0670}}$ \\ \hline
	\multirow{5}{*}{Strategy4} 
	 & All & 0.0447 $_{\pm \text{ 0.0199}}$ & \textbf{0.0311} $_{\pm \text{ 0.0276}}$  & 0.3453 $_{\pm \text{ 0.0366}}$ & 0.1892 $_{\pm \text{ 0.0447}}$ & 0.0577 $_{\pm \text{ 0.0285}}$ & 0.0474          $_{\pm \text{ 0.0277}}$ & 0.2402 $_{\pm \text{ 0.0733}}$ & 0.2828 $_{\pm \text{ 0.0279}}$  \\ 
	 & A   & \textbf{0.0614} $_{\pm \text{ 0.0312}}$ & 0.0726 $_{\pm \text{ 0.0525}}$  & 0.3284 $_{\pm \text{ 0.0539}}$ & 0.1216 $_{\pm \text{ 0.0774}}$ & 0.1458 $_{\pm \text{ 0.0614}}$ & 0.4315          $_{\pm \text{ 0.0648}}$ & 0.2453 $_{\pm \text{ 0.0753}}$ & 0.2852 $_{\pm \text{ 0.0444}}$  \\ 
	 & B   & 0.0872 $_{\pm \text{ 0.0375}}$ & \textbf{0.0770} $_{\pm \text{ 0.0416}}$ & 0.1858 $_{\pm \text{ 0.0770}}$ & 0.2006 $_{\pm \text{ 0.0640}}$ & 0.0934 $_{\pm \text{ 0.0741}}$ & 0.1307          $_{\pm \text{ 0.0825}}$ & 0.2420 $_{\pm \text{ 0.0766}}$ & 0.2426 $_{\pm \text{ 0.0927}}$  \\ 
	 & C   & 0.0392 $_{\pm \text{ 0.0185}}$ & \textbf{0.0204} $_{\pm \text{ 0.0129}}$ & 0.2488 $_{\pm \text{ 0.0234}}$ & 0.1744 $_{\pm \text{ 0.0611}}$ & 0.1120 $_{\pm \text{ 0.0923}}$ & 0.0441          $_{\pm \text{ 0.0262}}$ & 0.2424 $_{\pm \text{ 0.0748}}$ & 0.2583 $_{\pm \text{ 0.0863}}$   \\ 
	 & D   & 0.0512          $_{\pm \text{ 0.0408}}$ & 0.0579 $_{\pm \text{ 0.0508}}$ & 0.2849 $_{\pm \text{ 0.0384}}$ & 0.1985 $_{\pm \text{ 0.0330}}$ & 0.1266 $_{\pm \text{ 0.0465}}$ & \textbf{0.0365} $_{\pm \text{ 0.0213}}$  & 0.2446 $_{\pm \text{ 0.0739}}$ & 0.2783 $_{\pm \text{ 0.0925}}$   \\ \hline
	\multirow{5}{*}{Strategy5} 
	 & All & \textbf{0.0428} $_{\pm \text{ 0.0289}}$ & 0.0674 $_{\pm \text{ 0.0572}}$ & 0.2685          $_{\pm \text{ 0.0377}}$ & 0.1556 $_{\pm \text{ 0.0576}}$ & 0.0432 $_{\pm \text{ 0.0179}}$ & 0.0980 $_{\pm \text{ 0.0407}}$ & 0.2463 $_{\pm \text{ 0.0730}}$ & 0.2644 $_{\pm \text{ 0.0693}}$  \\ 
	 & A   & \textbf{0.0532} $_{\pm \text{ 0.0361}}$ & 0.0870 $_{\pm \text{ 0.0556}}$ & 0.3340          $_{\pm \text{ 0.0387}}$ & 0.0632 $_{\pm \text{ 0.0275}}$ & 0.1183 $_{\pm \text{ 0.1071}}$ & 0.0751 $_{\pm \text{ 0.0711}}$  & 0.2503 $_{\pm \text{ 0.0741}}$ & 0.2585 $_{\pm \text{ 0.0644}}$  \\ 
	 & B   & 0.3946          $_{\pm \text{ 0.1813}}$ & 0.3804 $_{\pm \text{ 0.2073}}$ & \textbf{0.2675} $_{\pm \text{ 0.2087}}$ & 0.2767 $_{\pm \text{ 0.1861}}$ & 0.3735 $_{\pm \text{ 0.2370}}$ & 1.0166 $_{\pm \text{ 0.3828}}$  & \textbf{0.2511} $_{\pm \text{ 0.0846}}$ & 0.2583 $_{\pm \text{ 0.1560}}$  \\ 
	 & C   & 0.1000 $_{\pm \text{ 0.0417}}$ & \textbf{0.0560} $_{\pm \text{ 0.0554}}$ & 0.1699          $_{\pm \text{ 0.0489}}$ & 0.1504 $_{\pm \text{ 0.0602}}$ & 0.1131 $_{\pm \text{ 0.0723}}$ & 0.1340 $_{\pm \text{ 0.0511}}$  & 0.2477 $_{\pm \text{ 0.0850}}$ & 0.2445 $_{\pm \text{ 0.1462}}$  \\ 
	 & D   & 0.1249 $_{\pm \text{ 0.1068}}$ & \textbf{0.1159} $_{\pm \text{ 0.0987}}$ & 0.2032          $_{\pm \text{ 0.0693}}$ & 0.2139 $_{\pm \text{ 0.1237}}$ & 0.3117 $_{\pm \text{ 0.1843}}$ & 0.1310 $_{\pm \text{ 0.0696}}$ & 0.2459 $_{\pm \text{ 0.0848}}$ & 0.2541 $_{\pm \text{ 0.1404}}$   \\ \hline
\end{tabular}
}
\end{center}
\end{table}

\begin{table}[!t] 
\renewcommand\arraystretch{1.05}
\caption{MAPE on Real-World dataset (All Surrogates) }
\vskip -0.1in
\label{tb:real-world result all surrogates}
\begin{center}
\scalebox{0.50}{
\begin{tabular}{cccccccccc} 
	& \ & LASER & LASER-GRU  & SInd-Linear   & SInd-MLP   & LTEE  & EETE & \red{BD-Linear} & \red{BD-MLP}      \\ \hline
	Strategy 
	 & City& mean             $_{\pm \text{ std}}$   & mean   $_{\pm \text{ std}}$    & mean   $_{\pm \text{ std}}$    & mean    $_{\pm \text{ std}}$   & mean $_{\pm \text{ std}}$  & mean $_{\pm \text{ std}}$  & \red{ mean $_{\pm \text{ std}}$}  & \red{mean $_{\pm \text{ std}}$} \\ \hline
	\multirow{5}{*}{Strategy1} 
	 & All & 0.0185 $_{\pm \text{ 0.0141}}$ & \textbf{0.0103}     $_{\pm \text{ 0.0100}}$     & 0.1814 $_{\pm \text{ 0.0126}}$  & 0.0962 $_{\pm \text{ 0.0142}}$ & 0.0256 $_{\pm \text{ 0.0155}}$ & 0.0244 $_{\pm \text{ 0.0148}}$  & 0.1804 $_{\pm \text{ 0.0142}}$  & 0.1398 $_{\pm \text{ 0.0183}}$  \\ 
	 & A   & \textbf{0.0146} $_{\pm \text{ 0.0188}}$ & 0.0179     $_{\pm \text{ 0.0136}}$     &  0.1384 $_{\pm \text{ 0.0463}}$ & 0.1028 $_{\pm \text{ 0.0525}}$ & 0.0449 $_{\pm \text{ 0.0423}}$ & 0.0692 $_{\pm \text{ 0.0266}}$ &  0.1591 $_{\pm \text{ 0.0398}}$  & 0.1292 $_{\pm \text{ 0.0242}}$  \\ 
	 & B   & 0.0861          $_{\pm \text{ 0.0510}}$ & \textbf{0.0718}     $_{\pm \text{ 0.0361}}$     &  0.1693 $_{\pm \text{ 0.0334}}$ & 0.1542 $_{\pm \text{ 0.0392}}$ & 0.0805 $_{\pm \text{ 0.0445}}$ & 0.2100 $_{\pm \text{ 0.0705}}$  &  0.1632 $_{\pm \text{ 0.0382}}$  & 0.1743 $_{\pm \text{ 0.0704}}$ \\ 
	 & C   & \textbf{0.0312} $_{\pm \text{ 0.0146}}$ & 0.0341     $_{\pm \text{ 0.0166}}$     &  0.1096 $_{\pm \text{ 0.0335}}$ & 0.0827 $_{\pm \text{ 0.0447}}$ & 0.0616 $_{\pm \text{ 0.0342}}$ & 0.0391 $_{\pm \text{ 0.0298}}$  &  0.1498 $_{\pm \text{ 0.0435}}$  & 0.1692 $_{\pm \text{ 0.0625}}$ \\ 
	 & D   & \textbf{0.0135} $_{\pm \text{ 0.0140}}$ & 0.0221     $_{\pm \text{ 0.0116}}$     &  0.0716 $_{\pm \text{ 0.0260}}$ & 0.0481 $_{\pm \text{ 0.0330}}$ & 0.0645 $_{\pm \text{ 0.0333}}$ & 0.0913 $_{\pm \text{ 0.0175}}$   &  0.1345 $_{\pm \text{ 0.0507}}$  & 0.1650 $_{\pm \text{ 0.0576}}$ \\ \hline
	\multirow{5}{*}{Strategy2} 
	 & All & \textbf{0.0263} $_{\pm \text{ 0.0157}}$ & 0.0295     $_{\pm \text{ 0.0172}}$     &  0.2159 $_{\pm \text{ 0.0126}}$ & 0.1281 $_{\pm \text{ 0.0207}}$ & 0.0334 $_{\pm \text{ 0.0176}}$ & 0.0634          $_{\pm \text{ 0.0189}}$ &  0.1481 $_{\pm \text{ 0.0555}}$  & 0.1622 $_{\pm \text{ 0.0307}}$ \\ 
	 & A   & \textbf{0.0123} $_{\pm \text{ 0.0125}}$ & 0.0151     $_{\pm \text{ 0.0113}}$     &  0.1127 $_{\pm \text{ 0.0224}}$ & 0.1030 $_{\pm \text{ 0.0110}}$ & 0.0484 $_{\pm \text{ 0.0322}}$ & 0.1109          $_{\pm \text{ 0.0311}}$   &  0.1431 $_{\pm \text{ 0.0535}}$   & 0.1409 $_{\pm \text{ 0.0592}}$ \\ 
	 & B   & 0.0449 $_{\pm \text{ 0.0407}}$ & \textbf{0.0408}     $_{\pm \text{ 0.0355}}$     &  0.2069 $_{\pm \text{ 0.0537}}$ & 0.2192 $_{\pm \text{ 0.0461}}$ & 0.1509 $_{\pm \text{ 0.1041}}$ & 0.0501 $_{\pm \text{ 0.0445}}$   &  0.1520 $_{\pm \text{ 0.0580}}$ & 0.1806 $_{\pm \text{ 0.0801}}$ \\ 
	 & C   & 0.0493          $_{\pm \text{ 0.0298}}$ & 0.0522     $_{\pm \text{ 0.0165}}$     &  0.1361 $_{\pm \text{ 0.0362}}$ & 0.1096 $_{\pm \text{ 0.0401}}$ & 0.0338 $_{\pm \text{ 0.0295}}$ & \textbf{0.0231}         $_{\pm \text{ 0.0362}}$  &  0.1504 $_{\pm \text{ 0.0561}}$   & 0.1819 $_{\pm \text{ 0.0716}}$ \\ 
	 & D   & 0.0529 $_{\pm \text{ 0.0426}}$ & \textbf{0.0360}     $_{\pm \text{ 0.0259}}$     &  0.1261 $_{\pm \text{ 0.0207}}$ & 0.0759 $_{\pm \text{ 0.0371}}$ & 0.0719 $_{\pm \text{ 0.0601}}$ & 0.0708         $_{\pm \text{ 0.0333}}$  &  0.1482 $_{\pm \text{ 0.0541}}$   & 0.1843 $_{\pm \text{ 0.0717}}$ \\ \hline
	\multirow{5}{*}{Strategy3} 
	 & All & \textbf{0.0146} $_{\pm \text{ 0.0085}}$ & 0.0194     $_{\pm \text{ 0.0132}}$     &  0.2704 $_{\pm \text{ 0.0148}}$ & 0.1361 $_{\pm \text{ 0.0227}}$ & 0.0388 $_{\pm \text{ 0.0258}}$ & 0.1758 $_{\pm \text{ 0.0365}}$  &  0.1604 $_{\pm \text{ 0.0646}}$   & 0.2137 $_{\pm \text{ 0.0219}}$\\ 
	 & A   & \textbf{0.0397} $_{\pm \text{ 0.0402}}$ & 0.0715     $_{\pm \text{ 0.0395}}$     &  0.1716 $_{\pm \text{ 0.0374}}$ & 0.1283 $_{\pm \text{ 0.0181}}$ & 0.0626 $_{\pm \text{ 0.0155}}$ & 0.2817 $_{\pm \text{ 0.0557}}$   &  0.1615 $_{\pm \text{ 0.0628}}$  & 0.1794 $_{\pm \text{ 0.0650}}$\\ 
	 & B   & \textbf{0.0358} $_{\pm \text{ 0.0281}}$ & 0.0834     $_{\pm \text{ 0.0414}}$     &  0.2279 $_{\pm \text{ 0.0378}}$ & 0.2180 $_{\pm \text{ 0.0466}}$ & 0.0978 $_{\pm \text{ 0.0738}}$ & 0.3576 $_{\pm \text{ 0.1009}}$  &  0.1666 $_{\pm \text{ 0.0637}}$  & 0.2409 $_{\pm \text{ 0.1183}}$\\ 
	 & C   & \textbf{0.0264} $_{\pm \text{ 0.0155}}$ & 0.0512     $_{\pm \text{ 0.0273}}$     &  0.1450 $_{\pm \text{ 0.0404}}$ & 0.0932 $_{\pm \text{ 0.0661}}$ & 0.0776 $_{\pm \text{ 0.0269}}$ & 0.1657 $_{\pm \text{ 0.0724}}$  &  0.1650 $_{\pm \text{ 0.0626}}$  & 0.2426 $_{\pm \text{ 0.1082}}$\\ 
	 & D   & 0.0584 $_{\pm \text{ 0.0324}}$ & \textbf{0.0315}     $_{\pm \text{ 0.0301}}$     &  0.0799 $_{\pm \text{ 0.0563}}$ & 0.0652 $_{\pm \text{ 0.0635}}$ & 0.0636 $_{\pm \text{ 0.041}}$ & 0.1083 $_{\pm \text{ 0.0376}}$   &  0.1593 $_{\pm \text{ 0.0659}}$  & 0.2265 $_{\pm \text{ 0.1072}}$ \\ \hline
	\multirow{5}{*}{Strategy4} 
	 & All & 0.0419 $_{\pm \text{ 0.0298}}$ & \textbf{0.0312}     $_{\pm \text{ 0.0286}}$     &  0.2911 $_{\pm \text{ 0.0281}}$ & 0.1540 $_{\pm \text{ 0.0390}}$ & 0.0481 $_{\pm \text{ 0.0479}}$ & 0.0445          $_{\pm \text{ 0.0281}}$   &  0.1682 $_{\pm \text{ 0.0727}}$   & 0.2553 $_{\pm \text{ 0.0393}}$\\ 
	 & A   & \textbf{0.0447} $_{\pm \text{ 0.0219}}$ & 0.0688     $_{\pm \text{ 0.0407}}$     &  0.1962 $_{\pm \text{ 0.0609}}$ & 0.0922 $_{\pm \text{ 0.0643}}$ & 0.0629 $_{\pm \text{ 0.0252}}$ & 0.4162          $_{\pm \text{ 0.0740}}$   &  0.1699 $_{\pm \text{ 0.0724}}$   & 0.2252 $_{\pm \text{ 0.0948}}$\\  
	 & B   & 0.0998 $_{\pm \text{ 0.0562}}$ & \textbf{0.0791}     $_{\pm \text{ 0.0447}}$     &  0.2123 $_{\pm \text{ 0.0408}}$ & 0.1345 $_{\pm \text{ 0.0565}}$ & 0.1092 $_{\pm \text{ 0.0640}}$ & 0.1264          $_{\pm \text{ 0.0862}}$    &  0.1725 $_{\pm \text{ 0.0719}}$  & 0.2536 $_{\pm \text{ 0.1123}}$ \\  
	 & C   & \textbf{0.0331} $_{\pm \text{ 0.0165}}$ & 0.0353     $_{\pm \text{ 0.0262}}$     &  0.1566 $_{\pm \text{ 0.0356}}$ & 0.1433 $_{\pm \text{ 0.0216}}$ & 0.0939 $_{\pm \text{ 0.0697}}$ & 0.0494       $_{\pm \text{ 0.0341}}$    &  0.1715 $_{\pm \text{ 0.0706}}$   & 0.2468 $_{\pm \text{ 0.1073}}$\\  
	 & D   & 0.0451          $_{\pm \text{ 0.0159}}$ & 0.0452     $_{\pm \text{ 0.0343}}$     &  0.1975 $_{\pm \text{ 0.0418}}$ & 0.1144 $_{\pm \text{ 0.0385}}$ & 0.0767 $_{\pm \text{ 0.045}}$ & \textbf{0.0350} $_{\pm \text{ 0.0303}}$   &  0.1730 $_{\pm \text{ 0.0696}}$   & 0.2582 $_{\pm \text{ 0.1118}}$\\ \hline
	\multirow{5}{*}{Strategy5} 
	 & All & 0.0536          $_{\pm \text{ 0.0225}}$ & 0.0630     $_{\pm \text{ 0.0479}}$     &  0.2394          $_{\pm \text{ 0.0319}}$ & 0.1136 $_{\pm \text{ 0.0461}}$ & \textbf{0.0408} $_{\pm \text{ 0.0335}}$ & 0.0986 $_{\pm \text{ 0.0372}}$  &  0.1763          $_{\pm \text{ 0.0698}}$  & 0.1587 $_{\pm \text{ 0.0430}}$\\  
	 & A   & \textbf{0.0588} $_{\pm \text{ 0.0279}}$ & 0.0618     $_{\pm \text{ 0.0428}}$     &  0.2044          $_{\pm \text{ 0.0527}}$ & 0.0606 $_{\pm \text{ 0.0319}}$ & 0.1049 $_{\pm \text{ 0.0603}}$ & 0.0989 $_{\pm \text{ 0.0760}}$   &  0.1777          $_{\pm \text{ 0.0694}}$  & 0.1399 $_{\pm \text{ 0.1006}}$\\  
	 & B   & 0.2638 $_{\pm \text{ 0.1978}}$ & 0.4357     $_{\pm \text{ 0.2901}}$     &  0.3650 $_{\pm \text{ 0.1525}}$ & 0.3098 $_{\pm \text{ 0.2495}}$ & 0.2745 $_{\pm \text{ 0.2103}}$ & 1.0327 $_{\pm \text{ 0.4053}}$   &  0.1855 $_{\pm \text{ 0.0824}}$  & \textbf{0.1799} $_{\pm \text{ 0.2317}}$ \\  
	 & C   & \textbf{0.0413} $_{\pm \text{ 0.0205}}$ & 0.0696     $_{\pm \text{ 0.0453}}$     &  0.1013     $_{\pm \text{ 0.0404}}$ & 0.1538 $_{\pm \text{ 0.0598}}$ & 0.1126 $_{\pm \text{ 0.0489}}$ & 0.1377 $_{\pm \text{ 0.0576}}$  &  0.1821     $_{\pm \text{ 0.0828}}$   & 0.1843 $_{\pm \text{ 0.2010}}$\\  
	 & D   & 0.1032          $_{\pm \text{ 0.0669}}$ & \textbf{0.0567}     $_{\pm \text{ 0.0312}}$     &  0.1290          $_{\pm \text{ 0.0797}}$ & 0.0832 $_{\pm \text{ 0.0528}}$ & 0.1486 $_{\pm \text{ 0.1244}}$ & 0.1202 $_{\pm \text{ 0.0729}}$  &  0.1798    $_{\pm \text{ 0.0835}}$  & 0.1895 $_{\pm \text{ 0.1876}}$\\ \hline
\end{tabular}
}
\end{center}
\end{table}

We measure the effectiveness of our proposed methods using a real-world dataset collected from an online randomized controlled experiment that aims to estimate the long-term revenue increase of different taxi promotion strategies to users on the Chinese largest ride-hailing platform. The experiment is conducted in 4 different cities, and each enrolled user is randomly assigned one of the 6 different campaign strategies. In the next 90 days of the observation period, users' transaction histories are logged anonymously on the platform. To ensure the privacy of users, we sample the data in a stratum and aggregate them in batches. In the end, the observed samples resulted in a large panel dataset, which includes the first 14 days of intermediate revenue and 
users' behaviors that can be used as surrogates, as well as the primary outcome of interest, the $90$th day cumulative revenue per user. \red{To ensure the unconfoundedness assumption holds and the real long-term causal effects can be accessed, the observational and experimental data are all from this online randomized controlled experiment. Regarding the observational data, we use covariates, treatment, short-term outcome, and long-term outcome, i.e., $D_{obs} = \{ \langle x,t,m,y \rangle \}$. Regarding the experimental data, we only use covariates, treatment, and short-term outcome, i.e., $D_{exp} = \{ \langle x,t,m \rangle \}$.}

\red{Specifically, the original real-world dataset contains various short-term outcomes, including the user's short-term revenue increase, usage frequency, taxi empty rate and so on for each day in the first 14 days, and the long-term outcome in the dataset is the user's 90th-day revenue increase. We denote the whole original dataset as the dataset (All Surrogates) where we do not filter any short-term outcomes, resulting in 105 dimensions, and we denote the selective dataset as the dataset (Revenues Only) where only the first 14-day revenue increases are selected as the short-term outcomes due to the same practical meaning as the long-term outcome, which are of 14 dimensions. The covariates, the treatment, and the long-term outcome are identical in both datasets.}

For compatibility and simplicity with other methods, we convert campaign strategies into binary ones, by setting one baseline strategy as the control and the rest as treated. To further prove LASER can handle heterogeneity across different regions, we compare experiment results in total and 4 cities respectively. We list the detailed information of the real-world dataset in Table \ref{tb:detailed real dataset}, including the dimensions of covariates, short-term outcomes, etc. The detailed results are listed in Table \ref{tb:real-world result Benefits Only} and Table \ref{tb:real-world result all surrogates}.

In order to reflect the superior performance of the proposed method, we analyze the experiments by answering the following questions. 

\textbf{Which short-term outcomes should be used as surrogates?} Compared Table \ref{tb:real-world result all surrogates} with Table \ref{tb:real-world result Benefits Only}, the results of \textit{using all surrogates} are slightly better than the results of \textit{Revenues only}. Although the revenues are  the most important short-term outcome, adding more observational data will benefit the effectiveness of predicting the long-term outcome. To learn latent surrogate representation better, we suggest that we collect as many as possible short-term outcomes to achieve a better result and in a limited condition, we suggest collecting those surrogates which have the same practical meaning as the long-term outcome.

\textbf{Do we need to consider the temporal property of the short-term outcomes?} Compared LASER with LASER-GRU in Table \ref{tb:real-world result Benefits Only} and \ref{tb:real-world result all surrogates}, we find that these two methods perform almost the same. Thus we conclude that in the long-term causal effects estimating task, the time-dependent relationship is not important for long-term outcome estimation, because we are only concerned about the outcome at a fixed time step, i.e., $90$th day in our dataset, different from the traditional time series prediction task that may predict $90, 91, 92,...$ days' outcomes. That is also why LASER outperforms LTEE which focuses on capturing time-dependent relations to predict long-term outcomes.

\textbf{Is it necessary to conduct latent surrogate representation learning?} As shown in Table \ref{tb:real-world result Benefits Only} and \ref{tb:real-world result all surrogates}, our method generally performs better than other methods, which verifies Theorem \ref{theorem:ce}. It indicates that the representation learning method based on iVAE is valid for long-term causal effect estimating tasks. In detail, compared with SInd-Linear, SInd-MLP, \red{BD-Linear, BD-MLP} and EETE, our methods perform better in almost every strategy and city, which means in the real-world scenario the surrogacy assumption is usually violated indeed and it is necessary to learn the latent surrogate representation through their proxies.

\subsection{Semi-synthetic Datasets}
\label{sbsection: semi-synthetic}

\begin{figure}[!t]
\vskip 0.2in
\begin{center}
\subfigure[MAPE on the first semi-synthetic dataset.]{ \label{figure 4a}
    \includegraphics[width=0.45\textwidth]{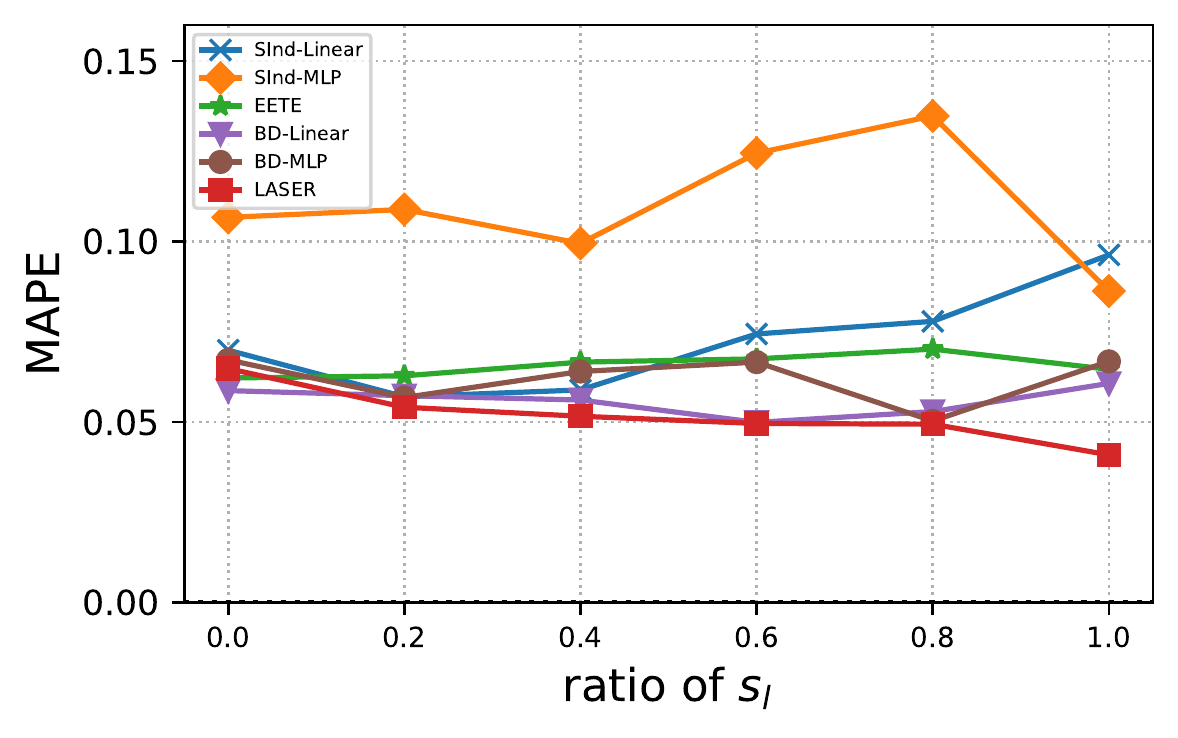}
}
\subfigure[MAPE on the \red{second} semi-synthetic dataset.]{ \label{figure 4b}
    \includegraphics[width=0.45\textwidth]{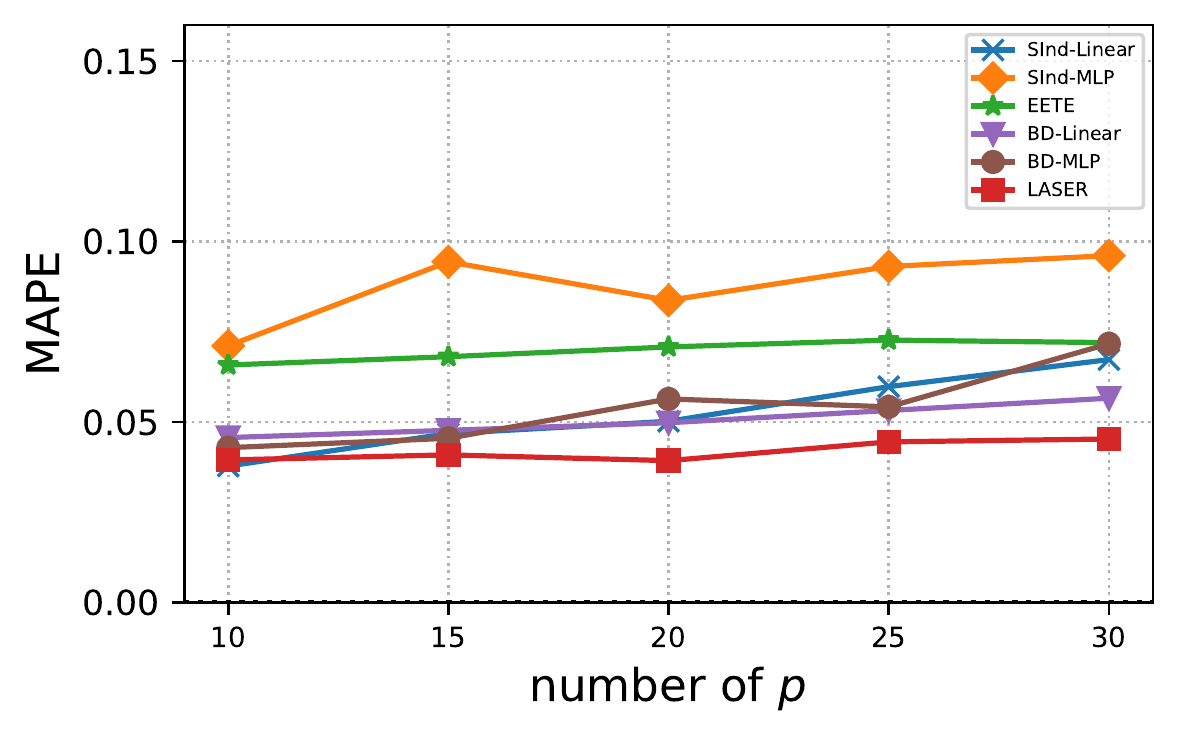}
}
\vskip -0.1in
\caption{MAPE on the semi-synthetic dataset IHDP.}
\label{figure:simulated result}
\end{center}
\vskip -0.2in
\end{figure}

Following LTEE \cite{cheng2021long}, we use the semi-synthetic dataset generated based on a benchmark dataset for causal inference -- IHDP \cite{hill2011bayesian}. The IHDP dataset is from a real-world randomized controlled trial,  designed to recover the effect of high-quality child care and home visits on the children's cognitive test scores. To mimic the reality, the imbalance between the treated and control units has been artificially introduced by removing a subset of the treated population and the outcomes are simulated using the original covariates and treatment \cite{hill2011bayesian}. Overall, this dataset contains 747 units and 25 covariates that describe both the characteristics of the infants and the characteristics of their mothers. To design a long-term setting dataset, we reuse the treatment and covariates in the IHDP dataset, and generate the surrogate and long-term outcome as follows:
\begin{equation*}
    \begin{aligned}
        & \begin{cases}
            s_{o}(1)= x (W_0 + \mathbf{1})  +\varepsilon _{0},\\
            s_{o}(0)= x (W_1 - \mathbf{1})  +\varepsilon _{0},
         \end{cases}
    \end{aligned}
\end{equation*}
\begin{equation*}
    \begin{aligned}
         & \begin{cases}
            s_{l}(1)= x (W_2 + \mathbf{1})  +\varepsilon _{1},\\
            s_{l}(0)= x (W_3 - \mathbf{1})  +\varepsilon _{1},
         \end{cases} 
    \end{aligned}
\end{equation*}
\begin{equation*}
    \begin{aligned}
        &  p = s_{l} W_{4}  + 5 \varepsilon _{2},  \\
        &  y = s_{l} W_{5} + s_{o} W_{6}  + x W_7 + \varepsilon _{3} ,  
    \end{aligned}
\end{equation*}
where  $W_j$ ($j=0,...,7$) are constant matrices or vectors and each of their elements is randomly sampled from $ \mathcal{N}(0,1)$, and $\varepsilon_j (j=0,1,2,3)$ are standard Gaussian. Specifically, $y$ is of dimensions $1$, and the coefficient of $\epsilon _{2}$ is $5$ to artificially make the proxies noisy. 

We generate two datasets to verify our method and theory. In the first dataset, we study the effect of the ratio of latent surrogates. Thus we fixed the number of observed short-term outcomes as $5$ and varied the ratio of latent surrogates. The range of numbers of variables in the first dataset are listed as follows: $num(s_l)=r*5$, $num(s_o)=(1-r)*5$ and $num(p)=r*5$, where $r=\{0, \frac{1}{5},\frac{2}{5},\frac{3}{5},\frac{3}{5},1\}$. The result is shown in Figure \ref{figure 4a}. In the second dataset, we study the effect of a varying number of proxies. Thus we do not generate $s_o$ and the numbers of $s_l$ \red{are} fixed as $5$, and the range of numbers of proxies is: $num(p)=\{10,15,20,25,30\}$. The result is shown in Figure \ref{figure 4b}.

\textbf{Sensitivity analysis of the ratio of latent surrogates.}
As shown in Figure \ref{figure 4a}, overall, LASER achieves the lowest MAPE. In detail, \red{When the ratio of $s_l$ to $s$ is small, the traditional surrogacy assumption approximately holds and SInd-Linear, BD-Linear, BD-MLP, and EETE have similar performance to LASER. However, a}s the ratio of $s_l$ to $s$ increases, the performance of SInd-Linear\red{, BD-Linear, BD-MLP} and EETE becomes worse in different degrees. SInd-Linear gets worse most obviously, indicating it is not robust to the violation of the surrogacy assumption. SInd-MLP is unstable, and the reason may be the unmatched regression. LASER performs stable and effective, which means LASER is robust in different cases and verifies \red{C}orollary \ref{corollary:special case}.

\textbf{Sensitivity analysis of the number of proxies. }
As shown in Figure \ref{figure 4b}, LASER outperforms all baselines. In detail, as the number of noisy proxies increases, the MAPE of LASER stays below $0.05$ while other methods perform worse. The reason is that with the number of noisy proxies increasing, the noises, which bias the accuracy of causal effects estimation, increase too. The result shows that we can learn the latent surrogate representation well with the noisy proxies, whenever the number of proxies is small or large.

\begin{figure}[htp]
\begin{center}
\subfigure[Visualization of $p(s|x,t)$ and $ \hat{p}(s|x,t,m)$]{   \label{fig:b}
    \includegraphics[width=0.8\textwidth]{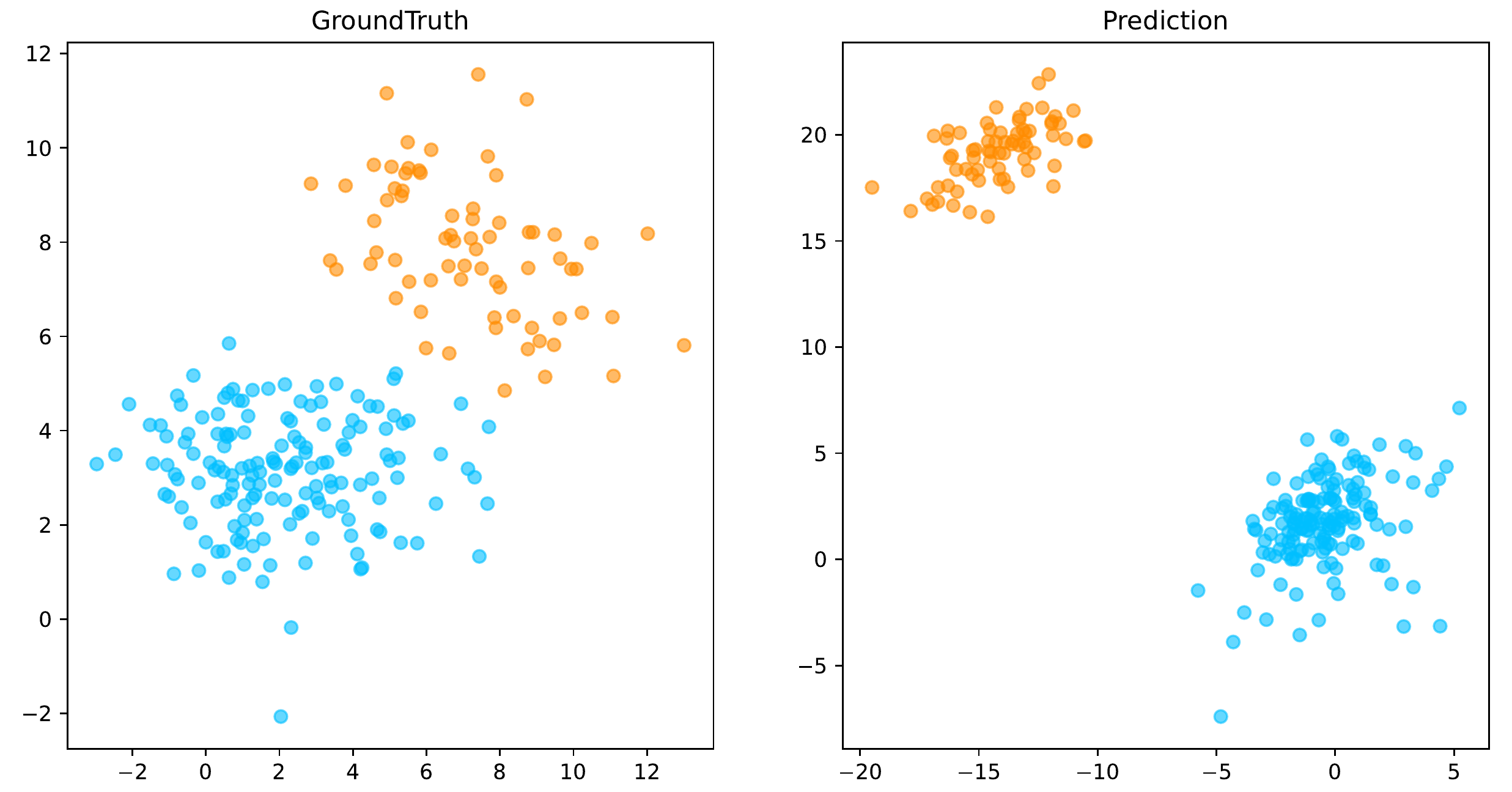}
}
\caption{\red{Experimental visualization of groundtruth latent surrogates (left) and learn latent surrogate representation (right).}} \label{fig: addition simulated dataset}
\end{center}
\end{figure}

\red{\subsection{Additional Results of Semi-synthetic Dataset}
\label{sbsection: semi-synthetic 2}
We further validate whether the iVAE-based LASER method can learn the latent surrogate representation as indicated by Theorem \ref{theorem:recover latent}. Following the data generation in Section \ref{sbsection: semi-synthetic}, we fixed the number of latent surrogates $s_l$ as 2, and fixed the number of observed surrogates $s_o$ as 0. The latent surrogates have two segments divided by the auxiliary treatment variables as shown in Figure \ref{fig: addition simulated dataset}. Then after applying our method, the recovered distribution (right) is very close to the true distribution (left) with some indeterminacy of scale which is reasonable due to the  $\sim_A -identifiable$ in our theoretical results. Therefore, we conclude that the latent surrogates can be identified by LASER.}

\section{Conclusions}
\label{sec:conclusion}

In this paper, we provide a practical solution to the real-world problem of estimating long-term causal effects in the case that the surrogates are partially or even totally unobserved. Note that the observation of noisy proxies of the latent surrogates helps to learn a predictive representation for long-term causal effects estimation, even though the proxies and partially observed surrogates are mixed together and indistinguishable.
Therefore, we propose an iVAE-based approach to learn a surrogate representation of the full short-term outcomes, without the need to distinguish the observed surrogates and proxies. Through such a unified view of the surrogates and the proxies of latent surrogates, we further devise an out-of-box, unbiased, and theoretically guaranteed solution to the long-term causal effects estimating problem. The extensive experiments, especially the real-world data experiments, validate the effectiveness of our method. A clear next step is to extend our work to robustly estimate long-term causal effects without the unconfoundedness assumption.

\section*{Acknowledgments}
This research was supported in part by National Key R\&D Program of China (2021ZD0111501), National Science Fund for Excellent Young Scholars (62122022), Natural Science Foundation of China (61876043, 61976052), the major key project of PCL (PCL2021A12).

\bibliographystyle{elsarticle-num} 
\bibliography{cas-refs}

\begin{thebibliography}{10}
\expandafter\ifx\csname url\endcsname\relax
  \def\url#1{\texttt{#1}}\fi
\expandafter\ifx\csname urlprefix\endcsname\relax\def\urlprefix{URL }\fi
\expandafter\ifx\csname href\endcsname\relax
  \def\href#1#2{#2} \def\path#1{#1}\fi

\bibitem{hohnhold2015focusing}
H.~Hohnhold, D.~O'Brien, D.~Tang, Focusing on the long-term: It's good for
  users and business, in: Proceedings of the 21th ACM SIGKDD International
  Conference on Knowledge Discovery and Data Mining, 2015, pp. 1849--1858.

\bibitem{fleming1994surrogate}
T.~R. Fleming, R.~L. Prentice, M.~S. Pepe, D.~Glidden, Surrogate and auxiliary
  endpoints in clinical trials, with potential applications in cancer and aids
  research, Statistics in medicine 13~(9) (1994) 955--968.

\bibitem{prentice1989surrogate}
R.~L. Prentice, Surrogate endpoints in clinical trials: definition and
  operational criteria, Statistics in medicine 8~(4) (1989) 431--440.

\bibitem{frangakis2002principal}
C.~E. Frangakis, D.~B. Rubin, Principal stratification in causal inference,
  Biometrics 58~(1) (2002) 21--29.

\bibitem{athey2019surrogate}
S.~Athey, R.~Chetty, G.~W. Imbens, H.~Kang, The surrogate index: Combining
  short-term proxies to estimate long-term treatment effects more rapidly and
  precisely, Tech. rep., National Bureau of Economic Research (2019).

\bibitem{cheng2021long}
L.~Cheng, R.~Guo, H.~Liu, Long-term effect estimation with surrogate
  representation, in: Proceedings of the 14th ACM International Conference on
  Web Search and Data Mining, 2021, pp. 274--282.

\bibitem{kallus2020role}
N.~Kallus, X.~Mao, On the role of surrogates in the efficient estimation of
  treatment effects with limited outcome data, arXiv preprint arXiv:2003.12408
  (2020).

\bibitem{pearl2009causal}
J.~Pearl, Causal inference in statistics: An overview, Statistics surveys 3
  (2009) 96--146.

\bibitem{rubin1974estimating}
D.~B. Rubin, Estimating causal effects of treatments in randomized and
  nonrandomized studies., Journal of educational Psychology 66~(5) (1974) 688.

\bibitem{IPWrosenbaum1983central}
P.~R. Rosenbaum, D.~B. Rubin, The central role of the propensity score in
  observational studies for causal effects, Biometrika 70~(1) (1983) 41--55.

\bibitem{IPWrosenbaum1987model}
P.~R. Rosenbaum, Model-based direct adjustment, Journal of the American
  Statistical Association 82~(398) (1987) 387--394.

\bibitem{johansson2016learning}
F.~Johansson, U.~Shalit, D.~Sontag, Learning representations for counterfactual
  inference, in: International conference on machine learning, PMLR, 2016, pp.
  3020--3029.

\bibitem{shalit2017estimating}
U.~Shalit, F.~D. Johansson, D.~Sontag, Estimating individual treatment effect:
  generalization bounds and algorithms, in: International Conference on Machine
  Learning, PMLR, 2017, pp. 3076--3085.

\bibitem{louizos2017causal}
C.~Louizos, U.~Shalit, J.~M. Mooij, D.~Sontag, R.~S. Zemel, M.~Welling, Causal
  effect inference with deep latent-variable models, in: NIPS, 2017.

\bibitem{zhang2021treatment}
W.~Zhang, L.~Liu, J.~Li, Treatment effect estimation with disentangled latent
  factors, in: Proceedings of the AAAI Conference on Artificial Intelligence,
  Vol.~35, 2021, pp. 10923--10930.

\bibitem{shi2021invariant}
C.~Shi, V.~Veitch, D.~M. Blei, Invariant representation learning for treatment
  effect estimation, in: Uncertainty in Artificial Intelligence, PMLR, 2021,
  pp. 1546--1555.

\bibitem{assaad2021counterfactual}
S.~Assaad, S.~Zeng, C.~Tao, S.~Datta, N.~Mehta, R.~Henao, F.~Li, L.~C. Duke,
  Counterfactual representation learning with balancing weights, in:
  International Conference on Artificial Intelligence and Statistics, PMLR,
  2021, pp. 1972--1980.

\bibitem{lauritzen2004discussion}
S.~L. Lauritzen, O.~O. Aalen, D.~B. Rubin, E.~Arjas, Discussion on causality
  [with reply], Scandinavian Journal of Statistics 31~(2) (2004) 189--201.

\bibitem{gilbert2008evaluating}
P.~B. Gilbert, M.~G. Hudgens, Evaluating candidate principal surrogate
  endpoints, Biometrics 64~(4) (2008) 1146--1154.

\bibitem{chen2007criteria}
H.~Chen, Z.~Geng, J.~Jia, Criteria for surrogate end points, Journal of the
  Royal Statistical Society: Series B (Statistical Methodology) 69~(5) (2007)
  919--932.

\bibitem{ju2010criteria}
C.~Ju, Z.~Geng, Criteria for surrogate end points based on causal
  distributions, Journal of the Royal Statistical Society: Series B
  (Statistical Methodology) 72~(1) (2010) 129--142.

\bibitem{yin2020novel}
Y.~Yin, L.~Liu, Z.~Geng, P.~Luo, Novel criteria to exclude the surrogate
  paradox and their optimalities, Scandinavian Journal of Statistics 47~(1)
  (2020) 84--103.

\bibitem{athey2020combining}
S.~Athey, R.~Chetty, G.~Imbens, Combining experimental and observational data
  to estimate treatment effects on long term outcomes, arXiv preprint
  arXiv:2006.09676 (2020).

\bibitem{imbens2022long}
G.~Imbens, N.~Kallus, X.~Mao, Y.~Wang, Long-term causal inference under
  persistent confounding via data combination, arXiv preprint arXiv:2202.07234
  (2022).

\bibitem{vae2014original}
D.~P. Kingma, M.~Welling, \href{http://arxiv.org/abs/1312.6114}{Auto-encoding
  variational bayes}, in: Y.~Bengio, Y.~LeCun (Eds.), 2nd International
  Conference on Learning Representations, {ICLR} 2014, Banff, AB, Canada, April
  14-16, 2014, Conference Track Proceedings, 2014.
\newline\urlprefix\url{http://arxiv.org/abs/1312.6114}

\bibitem{rezende2014stochastic}
D.~J. Rezende, S.~Mohamed, D.~Wierstra, Stochastic backpropagation and
  approximate inference in deep generative models, in: International conference
  on machine learning, PMLR, 2014, pp. 1278--1286.

\bibitem{higgins2016beta}
I.~Higgins, L.~Matthey, A.~Pal, C.~Burgess, X.~Glorot, M.~Botvinick,
  S.~Mohamed, A.~Lerchner, beta-vae: Learning basic visual concepts with a
  constrained variational framework (2016).

\bibitem{lopez2018information}
R.~Lopez, J.~Regier, M.~I. Jordan, N.~Yosef, Information constraints on
  auto-encoding variational bayes, in: NeurIPS, 2018.

\bibitem{pmlr-v139-rybkin21a}
O.~Rybkin, K.~Daniilidis, S.~Levine,
  \href{https://proceedings.mlr.press/v139/rybkin21a.html}{Simple and effective
  vae training with calibrated decoders}, in: M.~Meila, T.~Zhang (Eds.),
  Proceedings of the 38th International Conference on Machine Learning, Vol.
  139 of Proceedings of Machine Learning Research, PMLR, 2021, pp. 9179--9189.
\newline\urlprefix\url{https://proceedings.mlr.press/v139/rybkin21a.html}

\bibitem{WANG2023}
Q.~Wang, T.~P. Breckon,
  \href{https://www.sciencedirect.com/science/article/pii/S0893608023001697}{Generalized
  zero-shot domain adaptation via coupled conditional variational
  autoencoders}, Neural Networks (2023).
\newblock \href {https://doi.org/https://doi.org/10.1016/j.neunet.2023.03.033}
  {\path{doi:https://doi.org/10.1016/j.neunet.2023.03.033}}.
\newline\urlprefix\url{https://www.sciencedirect.com/science/article/pii/S0893608023001697}

\bibitem{khemakhem2020variational}
I.~Khemakhem, D.~Kingma, R.~Monti, A.~Hyvarinen, Variational autoencoders and
  nonlinear ica: A unifying framework, in: International Conference on
  Artificial Intelligence and Statistics, PMLR, 2020, pp. 2207--2217.

\bibitem{yao2021learning}
W.~Yao, Y.~Sun, A.~Ho, C.~Sun, K.~Zhang, Learning temporally causal latent
  processes from general temporal data, arXiv preprint arXiv:2110.05428 (2021).

\bibitem{pearl2009causality}
J.~Pearl, Causality, Cambridge university press, 2009.

\bibitem{hill2011bayesian}
J.~L. Hill, Bayesian nonparametric modeling for causal inference, Journal of
  Computational and Graphical Statistics 20~(1) (2011) 217--240.

\bibitem{pearl1993bayesian}
J.~Pearl, [bayesian analysis in expert systems]: comment: graphical models,
  causality and intervention, Statistical Science 8~(3) (1993) 266--269.

\end{thebibliography}

\appendix
\setcounter{theorem}{0}
\setcounter{equation}{0}

\setcounter{figure}{0}

\setcounter{section}{0}

\section{Proof of Theorem \ref{theorem:recover latent}} \label{appendix: proof of latent identifiability}
We now prove Theorem \ref{theorem:recover latent} that under mild assumptions, we can recover the latent up to some simper transformations. The proof is based on the proof of Theorem 1 in Khemakhem et al. \cite{khemakhem2020variational}, with the following differences:
\begin{itemize}
\item We use both $t$ and $x$ as auxiliary variables, and treat $s$ and $\varepsilon$ as latent variables;
\item We consider different causal mechanism of generating observed variables $m=f(s,\varepsilon)$.
\end{itemize}

\begin{proof}
Suppose we have two sets of parameters  $\theta = (\mathbf{T},\mathbf{f},\mathbf{\lambda})$ and  $\hat \theta = (\mathbf{\hat T}, \mathbf{\hat f}^{}, \mathbf{\hat \lambda})$ such that $p_\theta(m|x,t)=p_{\hat \theta}(m|x,t)$, then: 

\begin{equation}  \label{eq: change se to x}
  \begin{aligned}
    &  p_{\mathbf{T},\mathbf{f},\mathbf{\lambda}}(m|x,t)=p_{\mathbf{\hat T}, \mathbf{\hat f}, \mathbf{\hat \lambda}}(m|x,t) 
    \\
  \implies & 
     \int_{S}  \int_{E}p_{\mathbf{T},\mathbf{\lambda}}(s,\varepsilon|x,t)p_{\mathbf{f}}  (m|s,\varepsilon) dsd\varepsilon
    = \int_{S}\int_{E}p_{\mathbf{\hat T},\mathbf{\hat \lambda}}(s,\varepsilon|x,t)p_{\mathbf{\hat f}} (m|s,\varepsilon) dsd\varepsilon
    \\
 \implies &  
     \int_{\mathcal{M}}p_{\mathbf{T},\mathbf{\lambda}}(\mathbf{f}^{-1}(\overline{m})|x,t) \text{vol} J_{\mathbf{f}^{-1}}(\overline{m}) p_{\mathbf{f}}(m|\mathbf{f}^{-1}(\overline{m}) d\overline{m}
    \\ & =
    \int_{\mathcal{M}}p_{\mathbf{\hat T},\mathbf{\hat \lambda}}(\mathbf{\hat f}^{-1}(\overline{m})|x,t) \text{vol} J_{\mathbf{\hat f^{-1}}}(\overline{m}) p_{\mathbf{\hat f}}(m|\mathbf{\hat f}^{-1}(\overline{m}) d\overline{m}
\end{aligned}
\end{equation}

In Eq. \ref{eq: change se to x}, we denote the volume of a matrix $\mathbf{A}$ as $\text{vol}\mathbf{A} := \sqrt{\text{det}\mathbf{A}^T\mathbf{A}}$. $J$ denotes the Jacobian. We made the change of variable $\overline{m}=\mathbf f(s,\varepsilon)$ on the left-hand side and $\overline{m}=\mathbf{\hat f} (s,\varepsilon)$ on the right-hand side.

In determining function $\mathbf{f}$, i.e., $m=\mathbf f(s,\varepsilon)$, the term $p_{\mathbf f}(m|\mathbf f^{-1}(\overline{m}))$ in Eq.\ref{eq: change se to x} can be written as $p(m-\overline{m})$ and then is vanished. Thus, we have the following equation:
\begin{equation}  \label{eq: step I result}
\begin{aligned}
    p_{\mathbf{T},\mathbf{\lambda}}(\mathbf{f}^{-1}(m)|x,t) \text{vol} J_{\mathbf{f}^{-1}}(m) 
    =
   p_{\mathbf{\hat T},\mathbf{\hat \lambda}}(\mathbf{\hat f}^{-1}(m)|x,t) \text{vol} J_{\mathbf{\hat f^{-1}}}(m) .
\end{aligned}
\end{equation}

 By taking the logarithm on both side of Eq. \ref{eq: step I result} and replacing $p_{\mathbf{T}, \mathbf{\lambda}}$ by its expression from \red{Eq. \ref{exp dist definition} under Assumption \ref{ass: exp dist}}, we have:
\begin{equation} 
    \begin{aligned}
         \log \text{vol} J_{\mathbf{f}^{-1}}(m) + \Sigma_{i=1}^{2n}(  \log & Q_i(f_i^{-1}(m))- \log Z_i(x,t) \\
        &  + \Sigma_{j=1}^{k} T_{i,j}(f_i^{-1}(m)) \lambda_{i,j}(x,t) )
        \\ = 
         \log \text{vol} J_{\mathbf{\hat f}^{ -1}}(m) + \Sigma_{i=1}^{2n}(  \log &  \hat{Q}_i (\hat f_i^{ -1}(m)) - \log \hat {Z}_i (x,t) \\
        &  + \Sigma_{j=1}^{k} \hat{T}_{i,j}(\hat{f}_i^{ -1}(m)) \hat {\lambda}_{i,j}(x,t) ).
    \end{aligned}
\end{equation}
Let $2nk+1$ distinct points $(x_0,t_0), ..., (x_{2nk+1},t_{2nk+1})$ be the point provided by the condition (2) of Theorem, and define $\overline{\bm{\lambda}}(x,t)=\bm{\lambda}(x,t)-\bm{\lambda}(x_0,t_0)$. We evaluate the above equations at these points to obtain $2nk+1$ equations, and subtract the first equation from the remaining $2nk$ equations to obtain, for $l=1, ..., 2nk$:
\begin{equation} \label{eq: 2n eq}
\begin{aligned}
     \langle  \mathbf{T} (\mathbf{f}^{-1}(m)) & ,\overline{ \bm{\lambda}}(x_l,t_l) \rangle
    + \Sigma_i \log \frac{Z_i(x_0,t_0)}{Z_i(x_l,t_l)}
    \\ = &
        \langle  \mathbf{\hat T} (\mathbf{\hat f}^{-1}(m)) ,\overline{ \bm{\hat \lambda}}(x_l,t_l) \rangle
    + \Sigma_i \log \frac{\hat{Z}_i(x_0,t_0)}{\hat{Z}_i(x_l,t_l)}
\end{aligned}
\end{equation}
Let $\mathbf{L}$ be the matrix defined in condition (2) and $\mathbf{\hat L}$ similarly defined for $\bm{\hat \lambda}$ (note that $\mathbf{\hat L}$ is not necessarily invertible). Define $b_l= \Sigma_i \log \frac{\hat{Z}_i(x_0,t_0)Z_i(x_l,t_l)}{Z_i(x_0,t_0)\hat{Z}_i(x_l,t_l)}$ and $\mathbf{b}$ the vector of all $b_l$ for $l=1,..., 2nk$.

Then Eq.\ref{eq: 2n eq} can be rewritten in the matrix form:
\begin{equation*}
    \mathbf{L}^T \mathbf{T} (\mathbf{f}^{-1}(m)) =  \mathbf{\hat L}^{ T} \mathbf{\hat T}  (\mathbf{\hat f}^{ -1}(m))  + \mathbf{b}.
\end{equation*}
Then we multiply both sides of the above equation by $\bm{L}^{-T}$ to get:
\begin{equation} \label{eq: result step II}
\begin{aligned}
     \mathbf{T} (\mathbf{f}^{-1}(m)) =  \mathbf{A}\mathbf{\hat T}  (\mathbf{\hat f}^{ -1}(m))  + \mathbf{c},
\end{aligned}
\end{equation}
where $\mathbf{A}=\mathbf{L}^{-T}\mathbf{\hat L} $ and $\mathbf{c} = \mathbf L^{-T}\mathbf b$.

 We next show that $\mathbf{A}$ is invertible. By definition of $\mathbf{T}$ and according to condition (1), its Jacobian exists and is a $ 2nk \times 2n$ matrix of rank $2n$. This implies that the Jacobian of $\mathbf{\hat T} \circ \mathbf{\hat f}^{ -1}$ exists and is of rank $2n$ and so is $\mathbf{A}$. 

If $k=1$, $\mathbf A$ is invertible because $\mathbf A$ is $2n \times 2n $ matrix of rank $2n$.

If $k>1$, $\mathbf A$ is also invertible because:

Define $\overline{m}=\mathbf{f}^{-1} (m)$ and $\mathbf{T}_i(\overline{m}_i) = (T_{i,1}(\overline{m}_i), ..., T_{i,k}(\overline{m}_i))$. 

First, we prove, by contradiction, that for each $ i \in [1,...,2n]$, there exist k points  $\overline{m}_i^1, ..., \overline{m}_i^k$ such that for the distribution $p_{\mathbf{T},\bm{\lambda}}$ the family $(\mathbf{T}_i^{\prime}(\overline{m}_i^1), ..., \mathbf{T}_i^{\prime}(\overline{m}_i^k))$ are linearly independent. Suppose for any choice of $\overline{m}_i^1, ..., \overline{m}_i^k$, the family $(\mathbf{T}_i^{\prime}(\overline{m}_i^1), ..., \mathbf{T}_i^{\prime}(\overline{m}_i^k))$
is never linearly independent. This means that $\mathbf{T}_i^{\prime}(\mathbb{R})$ is included in a subspace of $\mathbb{R}^k$ of the dimension of most $k-1$. Let $\mathbf h$ be a non-zero vector that is orthogonal to $\mathbf{T}_i^{\prime}(\mathbb{R})$. Then for all $m \in \mathbb{R}$, we have $\langle \mathbf{T}_i^{\prime}(\overline{m}), \mathbf{h} \rangle = 0$. By integrating we find that $\langle \mathbf{T}_i (m), \mathbf{h} \rangle = const$. Since this is true for all $m \in \mathbb R$ and a $ \mathbf{h} \neq 0$, we conclude that the distribution is not strongly exponential, which contradicts the definition of distribution of $p_{\mathbf{T},\bm{\lambda}}$. So by contradiction, we conclude that there exist k points $\overline{m}_i^1, ..., \overline{m}_i^k$ such that $(\mathbf{T}_i^{\prime}(\overline{m}_i^1), ..., \mathbf{T}_i^{\prime}(\overline{m}_i^k))$ are linearly independent.

Second, we prove the invertibility of $\mathbf A$. Collect these points to $k$ vectors $(\overline m^1, ..., \overline m^k)$ and concatenate the $k$ Jacobians $J_\mathbf{T}(\overline{m}^l)$ evaluated at each of those vectors horizontally into the matrix $\mathbf{Q}=(J_\mathbf{T}(\overline{m}^1), ..., J_\mathbf{T}(\overline{m}^k))$ and similarly define $\mathbf{\hat Q}$ as the concatenation of Jacobians of $\mathbf{\hat T} (\mathbf{\hat f} ^{ -1} \circ \mathbf f(\overline{m}))$ evaluated at those points. Then the matrix $Q$ is invertible. By differentiating Eq. \ref{eq: result step II} for each $\overline{m}^l$, we have:
\begin{equation*}
    \mathbf{Q} = \mathbf{A} \mathbf{\hat Q}.  
\end{equation*}
The invertibility of $\mathbf{Q} $ implies the invertibility of $\mathbf{A} $ and $\mathbf{\hat Q} $, which completes the proof.
\end{proof}

\section{Proof of Theorem \ref{theorem:ce}.} \label{appendix: proof of ce identifiability}

\begin{proof}
  The IPW method allows us to estimate causal effects by
  \begin{equation}
    \begin{aligned}
      \tau_{exp} & = \mathbb{E}[y(1) - y(0)] \\
           & = \mathbb{E}[y \frac{t}{e(x)} - y \frac{1-t}{1-e(x)}].
    \end{aligned}
  \end{equation}
  For simplicity, we only discuss the case of $t=1$. We define the unbiased long-term outcome predictor $h(s_l, s_o, x)=\mathbb{E}[y|s_l,s_o,x]$, and in practice the predictor that we use is $h_p(\hat{s},\hat{\varepsilon},x) = \mathbb{E}[y|\hat s,\hat{\varepsilon},x]$. 
  We bridge the two predictors as follows:
  \begin{equation}
    \begin{aligned}
      \mathbb{E}[y \frac{t}{e(x)}]  
        & = \mathbb{E}[h(s_l, s_o, x) \frac{t}{e(x)}] \\
        & = \mathbb{E}[\mathbb{E}[y|s_l, s_o,x] \frac{t}{e(x)}] \\
        & = \mathbb{E}[\mathbb{E}[y|s,x] \frac{t}{e(x)}] \\
        & = \mathbb{E}[\mathbb{E}[y|s, \varepsilon, x] \frac{t}{e(x)}] \\
        & = \mathbb{E}[\mathbb{E}[y|\mathbf{ f}(s,\varepsilon), x] \frac{t}{e(x)}] \\
        & = \mathbb{E}[\mathbb{E}[y|\mathbf{\hat f}^{-1} \circ  \mathbf{ f}(s,\varepsilon), x] \frac{t}{e(x)}] \\
        & = \mathbb{E}[\mathbb{E}[y|\mathbf{\hat f}^{-1}(m), x] \frac{t}{e(x)}] \\
        & = \mathbb{E}[\mathbb{E}[y|\mathbf{\hat s},\hat{\varepsilon}, x] \frac{t}{e(x)}] \\
        & = \mathbb{E}[ h_p(\hat{s},\hat{\varepsilon},x) \frac{t}{e(x)}]
    \end{aligned}
  \end{equation}
  where the fourth equation holds due to $\varepsilon \perp\!\!\!\perp y | x,s$, and the fifth equation holds due to the invertibility of $\mathbf f$.
  
  Hence, the long-term causal effects are identifiable.
\end{proof}

\section{Proof of Corollary \ref{corollary:special case}.} \label{app: proof of special cases}

\begin{proof}
As shown in Figure \ref{figure 3}, this two cases are special cases of our generalized causal graph in Figure \ref{figure 1b}, and we discuss respectively:

\textbf{Special case 1:} in this case, we observe the whole valid surrogates. We can consider this case as that we observe noiseless proxies with the latent surrogates: $s=s_o$ and also $m=s_o$, which means the $\mathbf{f}$ defined in Eq. \ref{dgp} is in a linear form with coefficient $1$, i.e., $m=\mathbf{f}(s)=1 \times s$. Thus the parameters $(\mathbf{T},\mathbf{f},\mathbf{\lambda})$ is still $\sim_A -identifiable$ when Theorem \ref{theorem:recover latent} hold. In such a case we can still recover $s$. Thus, suppose Theorem \ref{theorem:ce} holds, the causal effects are also identifiable.

\textbf{Special case 2:} in this case, we only observe the proxies of surrogates. Without observing one of the surrogates, we consider $s=s_l$ and $m=p$, which develops into the standard latent identification problem as discussed in Khemakhem et~al. \cite{khemakhem2020variational}. Thus the parameters $(\mathbf{T},\mathbf{f},\mathbf{\lambda})$ is $\sim_A -identifiable$ when Theorem \ref{theorem:recover latent} holds and then the causal effects are also identifiable when Theorem \ref{theorem:ce} holds.
\end{proof}

\section{ELBO Derivation} \label{appendix: elbo}
Our object function contains $ELBO$ and a extra loss $\mathcal{L}_y$ minimizing predicting error of long-term outcome $y$. We give the $ELBO$ derivation as follows:
\begin{equation}
\begin{aligned}
 & \mathbb{E}_{q_{D_{obs} \cup D_{exp}}} [\log p(m|x,t)] \\
 =& \mathbb{E}_{q_{D_{obs} \cup D_{exp}}} [\mathbb{E}_{q(s|m,x,t)}[\log p(m|x,t)+\log p(s|m,x,t)  -\log p(s|m,x,t)]]\\
 =& \mathbb{E}_{q_{D_{obs} \cup D_{exp}}} [\mathbb{E}_{q(s|m,x,t)}[\log p(m,s|x,t)-\log p(s|m,x,t) +\log q( s|m,x,t)  \\ 
  &-\log q( s|m,x,t)]]\\
 =& \mathbb{E}_{q_{D_{obs} \cup D_{exp}}} [\log p(m,s|x,t)-\log q( s|m,x,t) ]  +KL(q(s|m,x,t)|p(s|m,x,t))\\
 =& ELBO+KL(q(s|m,x,t)|p(s|m,x,t))\\
 \geq &  ELBO\\
 =& \mathbb{E}_{q_{D_{obs} \cup D_{exp}}} [\mathbb{E}_{q(s|m,x,t)} [\log p(s|x,t)+\log p(m|s)-\log q(s|m,x,t)]],
\end{aligned}
\end{equation}
where $q_{D_{obs} \cup D_{exp}}$ is the empirical data distribution given by dataset $D_{obs} \cup D_{exp}$.

\section{Additional Experiments}

\begin{table}[h] 
\caption{Detailed Information of Real-world Dataset in additional experiments} \label{tb:detailed real dataset 2}
\begin{center}
\scalebox{0.85}{
\begin{tabular}{cc|cl|c}
\hline
Subject & & \red{Property} &: \red{ Number}   & \red{Exist In}     \\ 
\hline
 $x$ & &  \red{dimension} &: 123  & \red{$D_{obs}, D_{exp}$ }       \\ 
\hline
 $m$ (all surrogates) &&  \red{dimension} &: 122 & \red{$D_{obs}, D_{exp}$ } \\ 
\hline
$m$ (revenues only) &&  \red{dimension} &: 7  &\red{$D_{obs}, D_{exp}$ }        \\ 
\hline
\red{ $y$} & & \red{dimension} &: \red{1}  & \red{ $D_{obs}$ }      \\ 
\hline
\multirow{8}{*}{sample size of observational data} 
& Strategy1 & \red{sample size} &: 34866     & \red{ $D_{obs}$ }       \\
& Strategy2 & \red{sample size} &: 34992     & \red{ $D_{obs}$ }       \\ 
& Strategy3 & \red{sample size} &: 35010     & \red{ $D_{obs}$ }       \\ 
& Strategy4 & \red{sample size} &: 34929     & \red{ $D_{obs}$ }       \\ 
& Strategy5 & \red{sample size} &: 35010     & \red{ $D_{obs}$ }       \\ 
& Strategy6 & \red{sample size} &: 34893     & \red{ $D_{obs}$ }       \\ 
& Strategy7 & \red{sample size} &: 34920     & \red{ $D_{obs}$ }       \\ 
& Strategy8 & \red{sample size} &: 35046     & \red{ $D_{obs}$ }       \\ 
\hline
\multirow{8}{*}{sample size of experimental data} 
& Strategy1 & \red{sample size} &: 3874      & \red{ $D_{exp}$ }      \\
& Strategy2 & \red{sample size} &: 3888      & \red{ $D_{exp}$ }      \\ 
& Strategy3 & \red{sample size} &: 3890      & \red{ $D_{exp}$ }      \\ 
& Strategy4 & \red{sample size} &: 3881      & \red{ $D_{exp}$ }      \\ 
& Strategy5 & \red{sample size} &: 3890      & \red{ $D_{exp}$ }      \\ 
& Strategy6 & \red{sample size} &: 3877      & \red{ $D_{exp}$ }      \\ 
& Strategy7 & \red{sample size} &: 3880      & \red{ $D_{exp}$ }      \\ 
& Strategy8 & \red{sample size} &: 3894      & \red{ $D_{exp}$ }      \\ 
\hline
\end{tabular}
}
\end{center}
\end{table}
 
\subsection{Additional Real-world Dataset}
To further validate the effectiveness of LASER, we conduct additional experiments on another much larger real-world dataset than the dataset in Section \ref{sbsection: real-world}. The dataset is collected from an online randomized controlled experiment on the Chinese largest ride-hailing platform, aiming to measure the effects of 9 different campaign strategies across 30 cities for 14 days. We record the first 7 days' user behavior as surrogates and the 14th day cumulative revenue as the long-term outcome. We compare LASER with baseline methods on 8 different campaign strategies. The detailed information of this real-world dataset is listed in Table \ref{tb:detailed real dataset 2}. The detailed  results on this dataset are listed in Table \ref{tb:additional exp only revenues} and Table \ref{tb:additional exp all surrogates}. The results are similar to the results in Section \ref{sbsection: real-world}, strengthening our experimental conclusions. Moreover, LASER still performs well in the larger dataset, showing its robustness in the high dimensional case.

\begin{table}[h] 
\renewcommand\arraystretch{1.1}
\caption{MAPE on Real-World dataset (Revenues Only) }
\label{tb:additional exp only revenues}
\begin{center}
\scalebox{0.50}{
\begin{tabular}{ccccccccc} 
	&  LASER & LASER-GRU  & SInd-Linear   & SInd-MLP   & LTEE  & EETE & \red{ BD-Linear} & \red{BD-MLP }    \\ \hline
	Strategy & mean             $_{\pm \text{ std}}$   & mean   $_{\pm \text{ std}}$    & mean   $_{\pm \text{ std}}$    & mean    $_{\pm \text{ std}}$   & mean $_{\pm \text{ std}}$  & mean $_{\pm \text{ std}}$ & \red{mean $_{\pm \text{ std}}$}  & \red{ mean $_{\pm \text{ std}}$} \\ \hline
	Strategy1&  \textbf{ 0.0296} $_{\pm \text{ 0.0069 }}$ &    0.0660  $_{\pm \text{ 0.0439 }}$     & 0.2787 $_{\pm \text{ 0.0231 }}$  & 0.2597 $_{\pm \text{ 0.0351 }}$ & 0.1247 $_{\pm \text{ 0.0251 }}$ & 0.1668 $_{\pm \text{ 0.0499 }}$  & 0.3060 $_{\pm \text{ 0.0172 }}$ & 0.4028 $_{\pm \text{ 0.0999 }}$ \\  	  
	\hline
 
	Strategy2&   0.0132 $_{\pm \text{ 0.0026 }}$ &    \textbf{ 0.0066} $_{\pm \text{ 0.0028 }}$     & 0.1867 $_{\pm \text{ 0.0014 }}$  &  0.1931 $_{\pm \text{ 0.0057 }}$ & 0.0388 $_{\pm \text{ 0.0284 }}$ &  0.1825 $_{\pm \text{ 0.0099 }}$  & 0.2523 $_{\pm \text{ 0.0550 }}$ & 0.3163 $_{\pm \text{ 0.1147 }}$\\ 	  
	\hline
 
	Strategy3&  \textbf{ 0.0047} $_{\pm \text{ 0.0026 }}$ &     0.0086 $_{\pm \text{ 0.0019 }}$     &  0.1987 $_{\pm \text{ 0.0051}}$  & 0.1906 $_{\pm \text{ 0.0051 }}$ &  0.0281 $_{\pm \text{0.0172 }}$ &  0.1399 $_{\pm \text{ 0.0085 }}$  & 0.2415 $_{\pm \text{ 0.0475 }}$ & 0.1511 $_{\pm \text{ 0.0196 }}$\\ 	  
	\hline
 
	Strategy4&  \textbf{ 0.0047} $_{\pm \text{ 0.0031 }}$ &     0.0578 $_{\pm \text{ 0.0126 }}$     & 0.2032 $_{\pm \text{ 0.0051 }}$  &  0.1974 $_{\pm \text{ 0.0088 }}$ & 0.0151 $_{\pm \text{ 0.0135 }}$ &  0.0517 $_{\pm \text{ 0.0081 }}$  & 0.2370 $_{\pm \text{ 0.0419 }}$ & 0.1947 $_{\pm \text{ 0.0539 }}$\\ 	  
	\hline
 
	Strategy5&  \textbf{ 0.0039} $_{\pm \text{ 0.0031 }}$ &     0.0056 $_{\pm \text{ 0.0026 }}$     & 0.1617 $_{\pm \text{ 0.0028 }}$  & 0.1620 $_{\pm \text{ 0.0023 }}$ & 0.0092 $_{\pm \text{ 0.0056 }}$ &  0.0442 $_{\pm \text{ 0.0079 }}$  & 0.2266 $_{\pm \text{ 0.0428 }}$ & 0.1986 $_{\pm \text{ 0.0084 }}$\\ 	  
	\hline
 
	Strategy6&  0.1137 $_{\pm \text{ 0.0804 }}$ &      \textbf{ 0.1076} $_{\pm \text{ 0.0379 }}$     & 0.1571 $_{\pm \text{ 0.0339 }}$  &  0.1838 $_{\pm \text{ 0.0519 }}$ &  0.1202 $_{\pm \text{ 0.0489 }}$ & 0.1840 $_{\pm \text{ 0.0736 }}$  & 0.2152 $_{\pm \text{ 0.0473 }}$ & 0.2583 $_{\pm \text{ 0.1181 }}$\\ 	  
	\hline
 
	Strategy7&  \textbf{ 0.0639 } $_{\pm \text{ 0.0565 }}$ &   0.0814   $_{\pm \text{ 0.0560 }}$     & 0.1078 $_{\pm \text{ 0.0186 }}$  &  0.1193 $_{\pm \text{ 0.0168 }}$ & 0.1435 $_{\pm \text{ 0.0793 }}$ &  0.1163 $_{\pm \text{ 0.0614 }}$  & 0.2057 $_{\pm \text{ 0.0499 }}$ & 0.1562 $_{\pm \text{ 0.1048 }}$\\ 	  
	\hline
 
	Strategy8&  0.0530 $_{\pm \text{ 0.0222 }}$ &     \textbf{ 0.0253} $_{\pm \text{ 0.0149 }}$     &  0.1578 $_{\pm \text{ 0.0061 }}$  &  0.1691 $_{\pm \text{ 0.0187 }}$ &  0.0674 $_{\pm \text{ 0.0142}}$ &  0.2081 $_{\pm \text{ 0.0389 }}$  & 0.2168 $_{\pm \text{ 0.0552 }}$ & 0.1871 $_{\pm \text{ 0.0947 }}$\\ 	  
	\hline
 
	Average  &\textbf{ 0.0358} $_{\pm \text{ 0.0222 }}$ &      0.0449 $_{\pm \text{ 0.0216 }}$     & 0.1815 $_{\pm \text{ 0.0120 }}$  & 0.1844 $_{\pm \text{ 0.0181 }}$ &  0.0684 $_{\pm \text{ 0.0290 }}$ &  0.1367 $_{\pm \text{ 0.0323 }}$  & 0.2376 $_{\pm \text{ 0.0446 }}$ & 0.2331 $_{\pm \text{ 0.0767 }}$\\ 
	\hline
\end{tabular}
}
\end{center}
\end{table}

\begin{table}[h] 
\renewcommand\arraystretch{1.1}
\caption{MAPE on Real-World dataset (All Surrogates) }
\vskip -0.1in
\label{tb:additional exp all surrogates}
\begin{center}
\scalebox{0.50}{
\begin{tabular}{cccccccccc} 
	&  LASER & LASER-GRU  & SInd-Linear   & SInd-MLP   & LTEE  & EETE & \red{ BD-Linear} & \red{BD-MLP }    \\ \hline
	Strategy & mean             $_{\pm \text{ std}}$   & mean   $_{\pm \text{ std}}$    & mean   $_{\pm \text{ std}}$    & mean    $_{\pm \text{ std}}$   & mean $_{\pm \text{ std}}$  & mean $_{\pm \text{ std}}$ & \red{ mean $_{\pm \text{ std}}$} & \red{mean $_{\pm \text{ std}}$} \\ \hline
 
	Strategy1&  \textbf{ 0.0822} $_{\pm \text{ 0.0350 }}$ &     0.1120 $_{\pm \text{ 0.0695 }}$     &  0.3067 $_{\pm \text{ 0.0197 }}$  &  0.2906 $_{\pm \text{ 0.0250 }}$ &  0.1059 $_{\pm \text{ 0.0294 }}$ &  0.1653 $_{\pm \text{ 0.0499 }}$  &  0.2783 $_{\pm \text{ 0.0206 }}$   &  0.2148 $_{\pm \text{ 0.1176 }}$\\  	  \hline
 
	Strategy2&  \textbf{ 0.0042} $_{\pm \text{ 0.0026 }}$ &     0.0480 $_{\pm \text{ 0.0302 }}$     &  0.1993 $_{\pm \text{ 0.0018 }}$  &  0.2058 $_{\pm \text{ 0.0055 }}$ & 0.0321 $_{\pm \text{ 0.0216 }}$ &  0.1827 $_{\pm \text{ 0.0090 }}$  &  0.2322 $_{\pm \text{ 0.0483 }}$ &  0.1616 $_{\pm \text{ 0.1017 }}$\\ 	  \hline
 
	Strategy3&  \textbf{ 0.0272} $_{\pm \text{ 0.0147 }}$ &      0.0467 $_{\pm \text{ 0.0190 }}$     & 0.2198 $_{\pm \text{ 0.0038 }}$  &  0.2255 $_{\pm \text{ 0.0028 }}$ & 0.0288 $_{\pm \text{ 0.0148 }}$ &  0.1390 $_{\pm \text{ 0.0084 }}$  &  0.2211 $_{\pm \text{ 0.0426 }}$ &  0.1851 $_{\pm \text{ 0.0670 }}$\\ 	  \hline
 
	Strategy4&  \textbf{ 0.0083} $_{\pm \text{ 0.0063 }}$ &     0.0123 $_{\pm \text{ 0.0046 }}$     &  0.2235 $_{\pm \text{ 0.0038 }}$  &  0.2310 $_{\pm \text{ 0.0072 }}$ &  0.0157 $_{\pm \text{ 0.0107}}$ &  0.0515 $_{\pm \text{ 0.0084 }}$  &  0.2166 $_{\pm \text{ 0.0378 }}$ &  0.1964 $_{\pm \text{ 0.0495 }}$\\ 	  \hline
 
	Strategy5&  \textbf{ 0.0043} $_{\pm \text{ 0.0024 }}$ &     0.0066 $_{\pm \text{ 0.0028 }}$     &  0.1851 $_{\pm \text{ 0.0019 }}$  &  0.1941 $_{\pm \text{ 0.0037 }}$ &  0.0096 $_{\pm \text{ 0.0069 }}$ &  0.0444 $_{\pm \text{ 0.0075 }}$  &  0.2057 $_{\pm \text{ 0.0403 }}$ &  0.1518 $_{\pm \text{ 0.0385 }}$\\ 	  \hline
 
	Strategy6&  \textbf{ 0.0947} $_{\pm \text{ 0.0255 }}$ &    0.0986 $_{\pm \text{ 0.0507 }}$     & 0.1711 $_{\pm \text{ 0.0308 }}$  &  0.2188 $_{\pm \text{ 0.0298 }}$ & 0.1433 $_{\pm \text{ 0.0645 }}$ & 0.1853 $_{\pm \text{ 0.0716 }}$  &  0.1997 $_{\pm \text{ 0.0438 }}$ &  0.2254 $_{\pm \text{ 0.1334 }}$\\ 	  \hline
 
	Strategy7&  \textbf{ 0.0157} $_{\pm \text{ 0.0017 }}$ &      0.0206 $_{\pm \text{ 0.0146 }}$     &  0.1489 $_{\pm \text{ 0.0136 }}$  &  0.1322 $_{\pm \text{ 0.0103 }}$ & 0.1377 $_{\pm \text{ 0.0893 }}$ &  0.1194 $_{\pm \text{ 0.0647 }}$  &  0.1865 $_{\pm \text{ 0.0524 }}$ &  0.1706 $_{\pm \text{ 0.0843 }}$\\ 	  \hline
 
	Strategy8& 0.0314 $_{\pm \text{ 0.0196 }}$ &     \textbf{ 0.0146}  $_{\pm \text{ 0.0081 }}$     &  0.2952 $_{\pm \text{ 0.0080 }}$  & 0.2656 $_{\pm \text{ 0.0251 }}$ & 0.0603 $_{\pm \text{ 0.0065 }}$ &  0.2072 $_{\pm \text{ 0.0403 }}$  &  0.1828 $_{\pm \text{ 0.0500 }}$ &  0.1400 $_{\pm \text{ 0.0818 }}$\\ 	  \hline
 
	Average  &\textbf{ 0.0335} $_{\pm \text{ 0.0135 }}$ &     0.0449 $_{\pm \text{ 0.0249 }}$     & 0.2187 $_{\pm \text{ 0.0104 }}$  &  0.2205 $_{\pm \text{ 0.0137 }}$ &  0.0667 $_{\pm \text{ 0.0305 }}$ &  0.1369 $_{\pm \text{ 0.0325 }}$  &  0.2153 $_{\pm \text{ 0.0419 }}$ &  0.1807 $_{\pm \text{ 0.0842 }}$\\ 
	  \hline
\end{tabular}
}
\end{center}
\end{table}





\end{document}